\documentclass[10pt,twocolumn,letterpaper]{article}

\usepackage[pagenumbers]{cvpr}
\usepackage{multirow}

\usepackage{amsmath,amssymb}

\usepackage{xspace}

\usepackage{amsthm}

\definecolor{deltagreen}{rgb}{0.07,0.50,0.20}
\definecolor{deltared}{rgb}{0.80,0.12,0.12}
\newcommand{\dPos}[1]{\textcolor{deltagreen}{#1}}
\newcommand{\dNeg}[1]{\textcolor{deltared}{#1}}
\usepackage{algorithm}
\usepackage{algpseudocode}

\usepackage{booktabs}
\usepackage{placeins}
\usepackage{xcolor}

\definecolor{cvprblue}{rgb}{0.21,0.49,0.74}
\usepackage[pagebackref,breaklinks,colorlinks,allcolors=cvprblue]{hyperref}

\def\confName{CVPR}
\def\confYear{2027}

\title{Beyond Hard Tokens: Predictive Memory for Multimodal Protein Generation}

\author{
\parbox{0.92\textwidth}{\centering
  Xiaoliang~Shi$^{1}$, Zichen~Wang$^{1}$, Zhijian~Wei$^{1}$, Zhanpeng~Shi$^{2,3}$,\\
  Zhenghong~Zhou$^{1}$, Yunpeng~Wang$^{1}$, Shuangjia~Zheng$^{1\dagger}$\\[0.6em]
  $^{1}$Shanghai Jiao Tong University\\
  $^{2}$Fudan University\\
  $^{3}$Shanghai Innovation Institute
}}

\begin{document}
\maketitle
\begingroup\renewcommand\thefootnote{\fnsymbol{footnote}}\footnotetext[2]{Corresponding author: shuangjia.zheng@sjtu.edu.cn}\endgroup

\begin{abstract}
Multimodal protein language models unify sequence and discrete structural tokens via iterative masked decoding, offering a promising paradigm for joint protein design. However, standard decoders rely on a memoryless mask reset at each iteration: only a sparse subset of newly committed hard tokens is carried forward, while all remaining unresolved sites are reset to an identical \texttt{[MASK]} embedding. In open-ended co-generation where neither modality provides a fixed external anchor, this stateless transition limits single-step refinement depth and deprives both tracks of stable cross-step soft anchors, leading to slow convergence or degenerative repetitive collapse. Observing that uncommitted predictions evolve with pronounced temporal coherence rather than erratic noise, we introduce \emph{Predictive Memory} (PredMem), a training-free self-conditioning recurrence that carries these soft beliefs across steps. At each step, PredMem summarizes raw logits into a probability-weighted Top-$k$ expectation embedding at every residue and affinely interpolates it into eligible masked inputs on the next step at zero extra denoiser cost. Across DPLM-2, DPLM-2.1, and ESM3, PredMem raises unconditional sequence--structure co-generation success by 3.6, 17.7, and 48.7 percentage points, respectively, while improving the quality--diversity trade-off and suppressing repetitive mode collapse. Controlled diagnostics show that these gains stem primarily from early trajectory stabilization in unanchored cross-modal co-evolution rather than extra compute or simple entropy reduction, establishing PredMem as an efficient inference mechanism for multimodal discrete generation.
\end{abstract}

\section{Introduction}
\label{sec:intro}

\begin{figure*}[t]
    \centering
    \includegraphics[width=\textwidth]{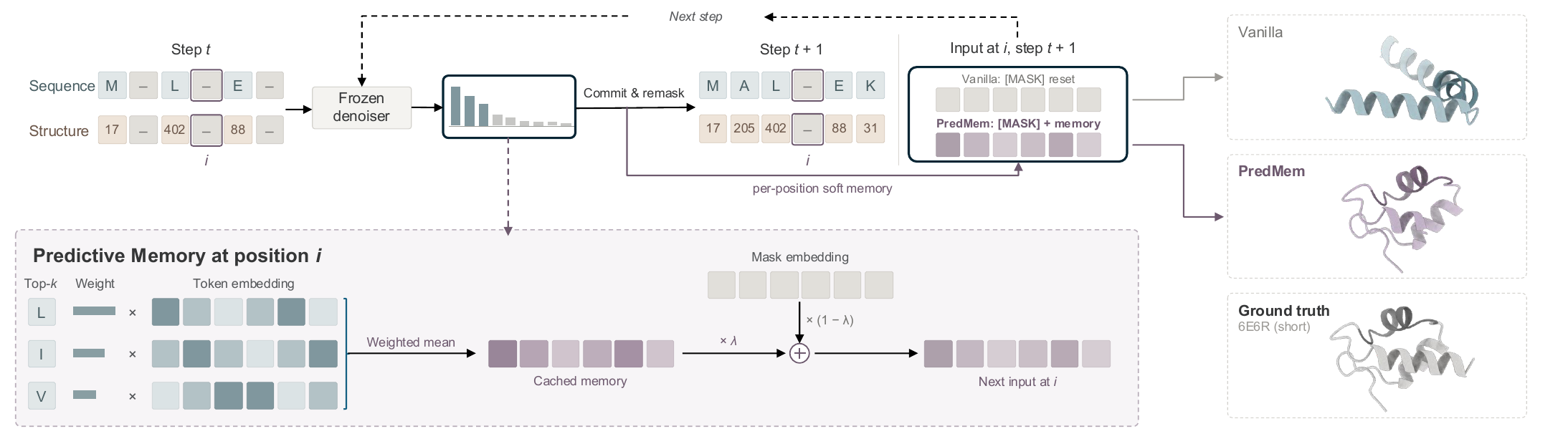}
    \caption{\textbf{Predictive Memory overview. }
PredMem passes each masked position a soft memory, a probability-weighted Top-$k$ token embedding interpolated with the mask embedding, instead of the vanilla \texttt{[MASK]} reset, with example motif-scaffolding outputs for 6E6R-short from vanilla decoding and PredMem, compared with the ground truth. }
    \label{fig:pm_overview}
\end{figure*}

Proteins are the molecular machinery of life, driving processes ranging from catalysis and immune defense to signal transduction~\cite{alberts2017molecular}. Their functions are governed by the sequence--structure--function relationship: the amino acid sequence encodes structural preferences that shape the three-dimensional conformation~\cite{anfinsen1973principles,jumper2021highly}, which in turn determines biological function. Computational protein design leverages this principle to engineer novel proteins with desired structural and functional properties~\cite{huang2016coming}.

Recent generative models have substantially advanced this goal, initially through unimodal approaches---including structure-based diffusion models such as RFdiffusion \cite{watson2023rfdiffusion}, Chroma \cite{ingraham2023chroma}, and FrameDiff \cite{yim2023framediff}, which generate backbones in continuous three-dimensional space, and protein language models such as ProtGPT2 \cite{ferruz2022protgpt2}, ProGen2 \cite{nijkamp2023progen2}, and ProGen3 \cite{bhatnagar2025progen3}, which generate amino acid sequences. However, sequences and structures are intrinsically coupled through shared evolutionary and biophysical constraints~\cite{rives2021biological,lin2023evolutionary}, so modeling either modality in isolation limits the ability to capture their interdependencies. Multimodal generative models, exemplified by ESM3 \cite{hayes2025simulating} and the DPLM-2 series \cite{wang2025dplm2,hsieh2025designspace}, address this by representing three-dimensional conformations as discrete structural tokens and modeling both modalities in a unified token space. Through iterative masked decoding, these models predict and commit tokens across both tracks, aiming to generate mutually self-consistent protein sequences and structures.

A foundational criterion for joint sequence--structure generation is biophysical self-consistency: whether a structure independently predicted from the generated sequence agrees with the co-generated backbone~\cite{trippe2023diffusion,yim2023framediff,hayes2025simulating,wang2025dplm2}. Yet maintaining this self-consistency remains a key bottleneck for existing multimodal models when both tracks must be resolved jointly without a fixed counterpart, such as in unconditional co-generation and motif scaffolding~\cite{wang2022scaffolding,watson2023rfdiffusion,zhou2026unlocking}. We identify the \emph{memoryless mask reset}: at each step, the denoiser predicts full distributions over all unresolved sites, but the sampler propagates only a sparse subset of newly committed hard tokens; all remaining positions—whether left masked or re-masked—are reset to an identical \texttt{[MASK]} embedding~\cite{ghazvininejad2019maskpredict,chang2022maskgit,sahoo2024mdlm}. This stateless transition erases local context at re-masked sites and forces persistently masked sites to reconstruct continuous hypotheses from scratch within a fixed single-step forward depth, rather than refining them recurrently across steps. 
Unlike single-modality generation with a dense static anchor, early joint generation starts from sparse commitments on both tracks; without a cross-step continuous state to stabilize intermediate beliefs, neither track can act as a reliable soft anchor for the other, causing slow convergence or repetitive collapse~\cite{holtzman2020curious}.
This raises a central question: 
do uncommitted intermediate predictions 
evolve with sufficient temporal coherence 
to act as a recurrent self-conditioning state across steps?

Inspecting decoding trajectories shows that uncommitted predictions do not fluctuate as erratic noise; instead, they exhibit pronounced temporal coherence at each residue. Across consecutive steps, Top-1 predictions at still-masked positions agree more frequently than position-agnostic baselines. This smooth evolution suggests reusable position-specific information. Existing inference methods leave this unused: track-wise classifier-free guidance (CFG) \cite{zhou2026unlocking} reshapes logits within a single step while still resetting unresolved sites across iterations, whereas self-conditioning and soft-masking in discrete diffusion \cite{cardei2026simple,hersche2026soft} require architectural changes or retraining, precluding plug-and-play use on frozen models. We introduce \emph{Predictive Memory} (PredMem), a training-free self-conditioning mechanism operating purely at the denoiser's input--output interface to stabilize early co-generation. At each step, PredMem summarizes raw, unguided logits into a probability-weighted Top-$k$ token embedding at every position and blends it into the native \texttt{[MASK]} embedding at eligible unresolved sites, keeping fixed conditions and protected motifs untouched at zero extra denoiser cost. 

Evaluated across ESM3, DPLM-2, and DPLM-2.1, PredMem consistently boosts unconditional co-generation design success—from \textbf{29.8\% to 33.4\% on DPLM-2}, and most substantially from \textbf{28.9\% to 46.6\% on DPLM-2.1} and \textbf{33.4\% to 82.1\% on ESM3}. Controlled experiments confirm that position-shuffled memory markedly weakens these gains, while entropy-controlled and compute-matched baselines rule out uncertainty reduction or extra denoising steps as sole explanations. Moreover, benefits concentrate where both tracks co-evolve freely—joint co-generation and ESM3 motif scaffolding, showing that \textbf{carrying position-specific soft beliefs as a recurrent self-conditioning state} enhances sequence--structure self-consistency in unanchored multimodal generation.

Our main contributions are summarized as follows:
\begin{enumerate}
\item \textbf{Temporal Coherence of Uncommitted Beliefs:} We show that intermediate uncommitted predictions in masked decoding evolve smoothly across steps, providing a stable continuous signal for feedback into frozen denoisers without retraining.
\item \textbf{Predictive Memory:} We propose a training-free inference-time self-conditioning recurrence that blends probability-weighted Top-$k$ expectation embeddings into eligible \texttt{[MASK]} inputs at zero extra denoiser cost.
\item \textbf{Systematic Validation \& Mechanistic Insights:} On DPLM-2, DPLM-2.1, and ESM3, PredMem consistently improves co-generation success, while controlled diagnostics pinpoint early trajectory stabilization and unanchored cross-modal co-evolution as the primary drivers.
\end{enumerate}

\section{Related Work}
\label{sec:related_work}

\textbf{Multimodal protein generation.}
Discrete multimodal generators unify protein sequence and quantized structure tokens via masked or diffusion modeling, notably ESM3~\cite{hayes2025simulating}, DPLM-2~\cite{wang2025dplm2} (extending the sequence-only DPLM~\cite{wang2024diffusion}), and DPLM-2.1~\cite{hsieh2025designspace}, while multimodal flows couple continuous backbones with discrete sequences~\cite{campbell2024discreteflows} and all-atom diffusion co-designs backbone, sequence, and sidechains~\cite{chu2024allatom}.
Rather than modifying architectures, we examine how frozen discrete generators reuse intermediate predictions to maintain sequence--structure self-consistency during joint decoding.

\textbf{Iterative masked decoding and inference-time guidance.}
Absorbing-state discrete diffusion~\cite{austin2021structured,sahoo2024mdlm} and confidence-based parallel decoders such as Mask-Predict~\cite{ghazvininejad2019maskpredict} and MaskGIT~\cite{chang2022maskgit} iteratively commit tokens while resetting remaining masks, and remasking samplers~\cite{wang2026remasking} further let committed tokens be revised.
At inference time, Zhou et al.~\cite{zhou2026unlocking} adapt classifier-free guidance (CFG)~\cite{ho2022classifierfree} and reward-guided beam search to multimodal proteins. Guidance for discrete diffusion~\cite{nisonoff2025unlocking} and for protein sequence generators~\cite{gruver2023protein,xiong2025proteinguide} likewise steers each denoising step toward target properties.
Unlike methods that only steer within-step logits or select hard tokens, PredMem propagates soft beliefs for unresolved sites across steps, complementing track-wise CFG.

\textbf{Self-conditioning and soft masking.}
Self-conditioning feeds previous denoiser estimates into subsequent steps in continuous diffusion~\cite{chen2022analog,strudel2022self}, iterative prompting~\cite{zhou2024diffusionnat}, and adapted masked diffusion~\cite{cardei2026simple}.
Most closely related are soft masking~\cite{hersche2026soft} and DMax~\cite{chen2026dmax}, which blend previous top-$k$ or intermediate embeddings into masked inputs but require model training or adaptation.
In contrast, PredMem operates completely training-free on frozen multimodal protein generators, maintaining position-specific, branch-local caches across both tracks with zero extra denoiser passes.
\section{Predictive Memory}
\label{sec:pm_method}

To overcome the \emph{memoryless mask reset} of iterative masked decoding without
retraining, PredMem introduces a lightweight, position-specific cache of the
model's previous predictions. At each step, PredMem summarizes the top-$k$
candidate token embeddings at every residue and interpolates these summaries
into eligible \texttt{[MASK]} positions during the subsequent forward pass,
leaving the frozen generator, sampling rule, and guidance mechanics completely
intact. Below, we first establish the empirical motivation
(\Cref{sec:motivation}) before formalizing the recurrent state, read/write
operations, and guidance-compatible caching
(\Cref{sec:pm_setup}--\Cref{sec:pm_guidance}). Complete formulations are deferred to Appendix~\ref{app:pm_interfaces}.

\subsection{Temporal Coherence of Uncommitted Beliefs}
\label{sec:motivation}
\begin{figure}[t]
    \centering
    \includegraphics[width=\linewidth]{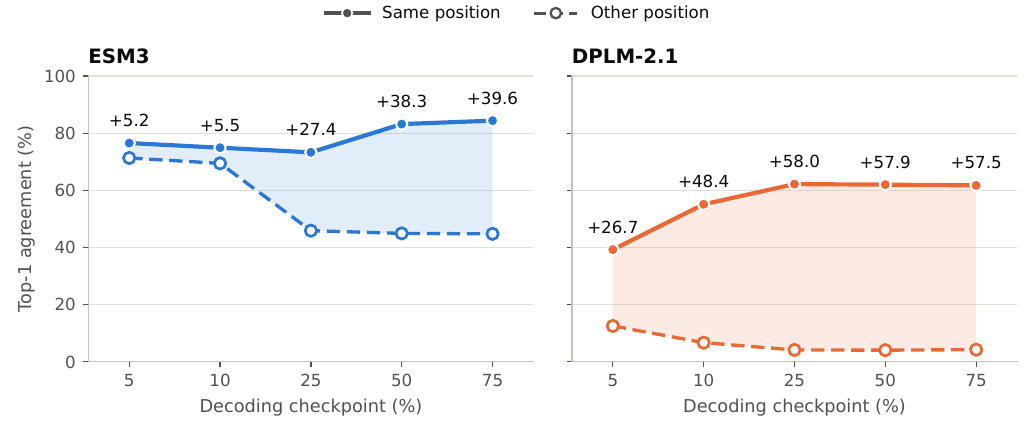}
    \caption{\textbf{Uncommitted predictions exhibit position-specific temporal coherence.}
 Top-1 agreement of adjacent-step predictions without PredMem, at the same position versus all other positions.}
    \label{fig:persist}
\end{figure}

Because the pretrained denoiser is frozen and never trained on soft
\texttt{[MASK]} inputs, feeding erratic or noisy intermediate predictions back
into the network could inject out-of-distribution perturbations and destabilize
decoding. We therefore examine whether uncommitted predictions evolve smoothly
across adjacent steps. Without PredMem, we take positions that are masked at two
adjacent steps and compare their Top-1 predictions at the same position with a
control that pairs each prediction with those at all other positions, which
reflects agreement due to token frequency alone (\Cref{fig:persist}; statistics
are computed per track and averaged over sequence and structure). Two
observations follow.

First, same-position agreement exceeds the control on both ESM3 and DPLM-2.1
throughout decoding, indicating that predictions at unresolved sites evolve
with position-specific temporal coherence rather than fluctuating randomly. This
motivates a recurrent cache indexed by position. Second, even when the discrete
Top-1 token changes, the consecutive tokens remain substantially closer in
embedding space than uniformly sampled distinct tokens, with cosine similarity
of $0.49$ on ESM3 and $0.53$ on DPLM-2.1, compared with $0.014$ and $0.097$
for uniform random distinct pairs. This local continuity in embedding space
motivates summarizing predictions as a soft expectation embedding rather than a
single hard token. The margin is smallest early in ESM3 decoding, where
predictions are highly repetitive and leave little headroom; yet carrying even
this modest position-specific asymmetry forward helps break global symmetry and
prevent early repetitive collapse.

\subsection{Iterative Masked Decoding and State}
\label{sec:pm_setup}

Let $z_t=(s_t,x_t)$ denote the partially masked sequence and structure tracks
at decoding step $t\in\{0,\ldots,T-1\}$ for a protein of length $L$, and let
$c$ denote fixed task conditions (e.g., a structural motif, a conditioning
sequence, or a fixed backbone). For each modality $m\in\{S,X\}$, let
$\mathcal V_m$ denote its valid token vocabulary and
$E_m:\mathcal V_m\cup\{\mathrm{MASK}\}\to\mathbb R^{d_m}$ its native token
embedding map.

At step $t$, the frozen denoiser $f_\theta$ is evaluated across one or more
decoding branches $b\in\mathcal G$ (where $\mathcal G=\{+\}$ for unguided
decoding and $\mathcal G=\{+,u_S,u_X\}$ under track-wise classifier-free
guidance; see \cref{sec:pm_guidance}). Each branch $b$ takes its
branch-specific input $(z_t^b,c^b)$ and outputs raw, unguided modality scores
$\ell_{t,i}^{b,m}(v)$ for every position $i$ and token $v\in\mathcal V_m$.

In standard iterative masked decoding~\citep{chang2022maskgit,sahoo2024mdlm}, any position $i$ that remains unresolved
at step $t$ is represented solely by the constant embedding
$E_m(\mathrm{MASK})$---the \emph{memoryless mask reset} that discards the
preceding predictive distribution $\ell_{t-1,i}^{b,m}$. PredMem instead
maintains a transient memory cache $M_{t-1}$ (initialized as
$M_{-1}=\varnothing$), where each entry $\mu_{t-1,i}^{b,m}\in M_{t-1}$ is
indexed by branch $b$, modality $m$, and residue position $i$. Because each
forward pass overwrites the cache with updated summaries $M_t$, past
predictions propagate recursively along the trajectory without storing
multi-step histories.

\subsection{Memory Write and Read}
\label{sec:pm_memory}

\paragraph{Writing predictive summaries.}
Immediately after the forward pass at step $t$, PredMem extracts a compact summary
from the raw scores $\ell_{t,i}^{b,m}$ prior to classifier-free guidance and
sampling-temperature scaling. Let $K_{t,i}^{b,m}\subseteq\mathcal V_m$ denote
the set of top-$k$ tokens under $\ell_{t,i}^{b,m}$ ($1\le k\le|\mathcal V_m|$).
We normalize the raw probabilities over this truncated support at unit
temperature:
\begin{equation}
  \widetilde p_{t,i}^{b,m}(v)
  =\frac{\exp\ell_{t,i}^{b,m}(v)}
  {\sum_{u\in K_{t,i}^{b,m}}\exp\ell_{t,i}^{b,m}(u)},
  \quad v\in K_{t,i}^{b,m},
  \label{eq:pm_topk}
\end{equation}
and write the resulting probability-weighted mean embedding into the cache:
\begin{equation}
  \mu_{t,i}^{b,m}
  =\sum_{v\in K_{t,i}^{b,m}}
  \widetilde p_{t,i}^{b,m}(v)E_m(v).
  \label{eq:pm_write}
\end{equation}
Computing the deterministic expectation $\mu_{t,i}^{b,m}$ at unit temperature
prior to guidance decouples cached beliefs from external sampling schedules,
keeps summaries within the convex hull of plausible top-$k$ embeddings, and
avoids injecting sampling variance into the next step. Summaries are written at both masked and
visible residue positions; thus, if a committed position is subsequently
remasked by the scheduler, its most recent predictive summary is immediately
available.

\paragraph{Reading at eligible masked positions.}
At step $t$, cached summaries from $M_{t-1}$ are read only into unresolved
positions designated for generation. Let $\mathcal I_m^{\mathrm{gen}}$ denote
the set of positions that the task allows the decoder to generate in modality
$m$, and let $\mathcal I_m^{\mathrm{fb}}\subseteq\mathcal I_m^{\mathrm{gen}}$
be the subset eligible for memory feedback (excluding fixed conditioning
residues, protected motif coordinates, and special BOS/EOS tokens). For an
active branch--modality pair $(b,m)\in\mathcal B$, the set of positions
receiving feedback at step $t$ is
\begin{equation}
  \mathcal A_t^{b,m}
  =\mathcal I_m^{\mathrm{fb}}
  \cap\{i:z_t^{b,m}(i)=\mathrm{MASK}\}.
  \label{eq:pm_eligible}
\end{equation}
When $M_{t-1}\ne\varnothing$, PredMem replaces the mask embedding at each active
position $i\in\mathcal A_t^{b,m}$ with the affine interpolation
\begin{equation}
  \widetilde e_{t,i}^{b,m}
  =(1-\lambda)E_m(\mathrm{MASK})
   +\lambda\mu_{t-1,i}^{b,m},
  \qquad i\in\mathcal A_t^{b,m},
  \label{eq:pm_read}
\end{equation}
where $\lambda\in[0,1]$ controls the feedback strength, while all remaining
positions $i\notin\mathcal A_t^{b,m}$ retain their native inputs. Through this
interpolation, each unresolved position carries a soft, position-specific
hypothesis across steps without prematurely committing to a hard token. Setting
$\lambda=0$ or $M_{-1}=\varnothing$ exactly recovers the baseline forward pass;
consequently, step $t=0$ remains identical to the unmodified baseline.

\paragraph{Architecture-specific instantiations.}
While Eq.~\eqref{eq:pm_read} defines the canonical read operation in token
space, different pretrained backbones preprocess embeddings differently prior to
self-attention. In ESM3~\citep{hayes2025simulating}, PredMem applies
Eq.~\eqref{eq:pm_read} directly to the sequence and structure token embeddings.
In DPLM-2 and DPLM-2.1~\citep{wang2025dplm2,hsieh2025designspace}, whose input layers apply mask-ratio rescaling followed
by LayerNorm, PredMem performs the interpolation at the post-LayerNorm attention
input after matching the native scaling factor. Furthermore, because DPLM-2.1
factorizes each structure token into 13 binary heads ($2^{13}=8192$ complete
codes), PredMem evaluates the joint log-probability across all $8192$ complete codes
to select the top-$k$ structural candidates and maps them through the native
codebook projection.
Details are in Appendix~\ref{app:pm_codes}.

\subsection{Guidance-Compatible Recurrence}
\label{sec:pm_guidance}

\paragraph{Branch-local memory under track-wise CFG.}
To steer multimodal generation without retraining, track-wise classifier-free
guidance~\citep{zhou2026unlocking} evaluates a fully conditioned branch $+$
alongside condition-dropped branches $u_m$, in which the conditioning input for
target modality $m$ is masked out. The guided
logits for modality $m$ are given by
\begin{equation}
  g_{t,i}^{m}
  =\ell_{t,i}^{u_m,m}
   +w_m\bigl(\ell_{t,i}^{+,m}-\ell_{t,i}^{u_m,m}\bigr),
  \label{eq:pm_cfg}
\end{equation}
where $w_m\ge 1$ is the guidance weight (with $w_m=1$ recovering the unguided
conditioned logits $\ell_{t,i}^{+,m}$).

Crucially, sharing a single cache across branches would leak conditioning
information from $+$ into $u_m$ or corrupt intentionally dropped tracks. PredMem
therefore maintains strictly branch-local caches. Specifically, in joint
generation with $\mathcal G=\{+,u_S,u_X\}$, only the generated target tracks of
each branch receive feedback, as defined by the active pair set
\begin{equation}
  \mathcal B=\{(+,S),(+,X),(u_S,S),(u_X,X)\}.
  \label{eq:pm_pairs}
\end{equation}
In single-modality guided generation, $\mathcal B$ reduces to
$\{(+,m),(u_m,m)\}$; in unguided generation ($\mathcal G=\{+\}$), $\mathcal B$
contains only the generated modalities of $+$. Each branch reads exclusively
from the cache constructed from its own unguided scores $\ell_{t-1}^{b,m}$,
ensuring that no memory is ever injected into a dropped condition.

\paragraph{Recurrent transition and computational cost.}
At each step $t$, the decoder reads from $M_{t-1}$ via Eq.~\eqref{eq:pm_read},
evaluates the frozen denoiser $f_\theta$ once across the active branches,
writes the updated cache $M_t$ via Eq.~\eqref{eq:pm_write}, and passes the
resulting guided scores $g_t$ to the unmodified native sampler. Because PredMem
operates purely at the denoiser's input--output interface, it preserves both
the total step count $T$ and the branch batch size $|\mathcal G|$: a trajectory
requires $T$ batched denoiser calls ($T|\mathcal G|$ branch evaluations) with or
without PredMem. The only computational overhead stems from top-$k$ selection,
embedding aggregation, and vector interpolation, which require
$O\!\left(L\sum_{(b,m)\in\mathcal B}d_m\right)$ memory independent of $T$.

\section{Experiments}
\label{sec:results}

\subsection{Experimental Setup}
\label{sec:exp-design}

\paragraph{Models and tasks.}
We evaluate three frozen multimodal protein generators: ESM3, DPLM-2, and
DPLM-2.1~\citep{hayes2025simulating,wang2025dplm2,hsieh2025designspace}.
Following \citet{zhou2026unlocking}, we consider two sequence--structure
co-generation tasks in the main text: unconditional co-generation and motif
scaffolding---while single-modality conditional tasks (structure
prediction and inverse folding~\citep{hsu2022inverse}) are deferred to the
appendix.

\paragraph{Controlled comparisons.}
For the two joint tasks, we compare
four configurations---Vanilla, PredMem, CFG, and CFG+PredMem---to evaluate
memory feedback with and without the inherited guidance.
Within each model and task, all four configurations share the decoding
schedule, step budget, position-selection rule, and sampling temperatures,
while CFG and CFG+PredMem retain the identical guidance settings from
\citet{zhou2026unlocking}. Contrasting PredMem against Vanilla and CFG+PredMem against
CFG therefore isolates memory feedback under 
matched denoiser-call and
branch-evaluation budgets.

\paragraph{Metrics and reporting.}
For joint tasks, ESMFold~\citep{lin2023evolutionary} independently refolds
generated sequences. We compare the refolded and generated structures using
backbone RMSD (scRMSD) and TM-score (scTM)~\citep{zhang2004tmscore}, and report refolded pLDDT as a confidence measure. These quality metrics are averaged over
all generated outputs, whereas Foldseek~\citep{van2024fast}
clustering (at a TM-score threshold of $0.5$, within each length or motif
variant) is computed strictly over the successful subset. Cluster counts
measure the breadth of the successful subset and depend on both yield and
diversity. Reported values are five-seed means; standard deviations, exact
hyperparameter settings, ablations on $\lambda$ and $k$, and single-modality
results are provided in Appendices~\ref{app:exp_protocol} and~\ref{app:full_results}.

\subsection{Unconditional Co-generation}
\label{sec:cogen}

\paragraph{Protocol.}
Both tracks start fully masked, without external conditioning. Each run
generates 500 pairs: 100 at each length in $\{100,200,300,400,500\}$.
A sample succeeds when $\mathrm{scRMSD}<2$~\AA.
\Cref{tab:main} reports success, structural quality, and successful clusters.

\definecolor{dposc}{RGB}{0,128,90}
\definecolor{dnegc}{RGB}{200,45,45}
\definecolor{dzeroc}{gray}{0.55}
\newcommand{\best}[1]{\textbf{#1}}
\newcommand{\second}[1]{\underline{#1}}
\newcommand{\dZero}[1]{\textcolor{dzeroc}{#1}}

\begin{table*}[t]
\centering
\small
\setlength{\tabcolsep}{4.5pt}
\renewcommand{\arraystretch}{1.06}
\caption{
\textbf{Unconditional co-generation.} Each run generates 500 sequence--structure pairs per model, measured over five seeds. 
Colored numbers show the change of PredMem over Vanilla and of CFG+PredMem over CFG. 
PredMem raises mean success on all three generators, most strongly on ESM3.
}
\label{tab:main}
\begin{tabular}{@{}ll*{5}{r@{\hspace{2pt}}l}@{}}
\toprule
& & \multicolumn{8}{c}{Quality} & \multicolumn{2}{c}{Diversity} \\
\cmidrule(lr){3-10} \cmidrule(l){11-12}
Model & Method
& \multicolumn{2}{c}{Success (\%) $\uparrow$}
& \multicolumn{2}{c}{pLDDT $\uparrow$}
& \multicolumn{2}{c}{scTM $\uparrow$}
& \multicolumn{2}{c}{scRMSD $\downarrow$}
& \multicolumn{2}{c}{\#Clusters $\uparrow$} \\
\midrule
\multirow{4}{*}{DPLM-2}
 & Vanilla & 29.80 &  & 82.00 &  & 0.852 &  & 4.95 &  & \second{78.0} & \\
 & PredMem & \second{33.40} & \dPos{+3.60} & \second{82.47} & \dPos{+0.47} & 0.852 & \dZero{+0.000} & 4.79 & \dPos{-0.16} & \best{81.0} & \dPos{+3.0} \\
 & CFG & 31.88 &  & 82.42 &  & \second{0.859} &  & \second{4.42} &  & 73.4 & \\
 & CFG+PredMem & \best{34.20} & \dPos{+2.32} & \best{82.99} & \dPos{+0.57} & \best{0.862} & \dPos{+0.003} & \best{4.37} & \dPos{-0.05} & 74.6 & \dPos{+1.2} \\
\midrule
\multirow{4}{*}{DPLM-2.1}
 & Vanilla & 28.88 &  & 84.91 &  & 0.836 &  & 5.17 &  & \second{92.6} & \\
 & PredMem & \second{46.60} & \dPos{+17.72} & \second{87.77} & \dPos{+2.86} & \second{0.889} & \dPos{+0.053} & \second{3.32} & \dPos{-1.85} & 75.6 & \dNeg{-17.0} \\
 & CFG & 32.88 &  & 85.14 &  & 0.853 &  & 4.37 &  & \best{98.0} & \\
 & CFG+PredMem & \best{48.32} & \dPos{+15.44} & \best{87.85} & \dPos{+2.71} & \best{0.895} & \dPos{+0.042} & \best{3.13} & \dPos{-1.24} & 68.4 & \dNeg{-29.6} \\
\midrule
\multirow{4}{*}{ESM3}
 & Vanilla & 33.36 &  & 67.91 &  & 0.545 &  & 31.91 &  & 62.0 & \\
 & PredMem & \best{82.08} & \dPos{+48.72} & \second{90.53} & \dPos{+22.62} & \best{0.944} & \dPos{+0.399} & \best{2.32} & \dPos{-29.59} & 73.0 & \dPos{+11.0} \\
 & CFG & 64.48 &  & 88.36 &  & 0.891 &  & 6.62 &  & \best{114.4} & \\
 & CFG+PredMem & \second{80.96} & \dPos{+16.48} & \best{91.59} & \dPos{+3.23} & \second{0.943} & \dPos{+0.052} & \second{2.47} & \dPos{-4.15} & \second{84.4} & \dNeg{-30.0} \\
\bottomrule
\end{tabular}
\end{table*}

\begin{figure}[t]
\centering
\includegraphics[width=\columnwidth,trim=12bp 68bp 19bp 33bp,clip]{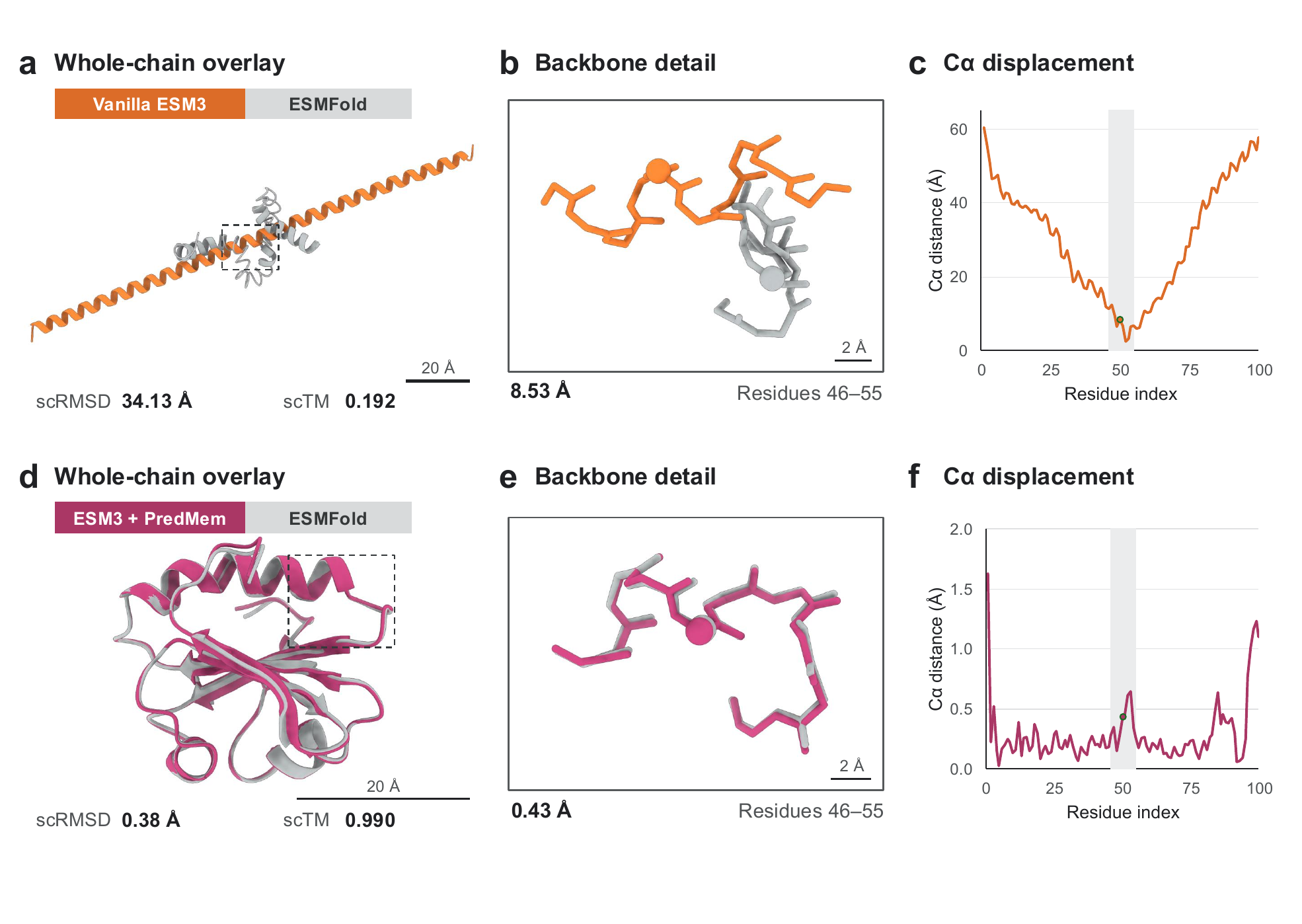}
\caption{\textbf{PredMem prevents repetitive collapse in ESM3.}
Vanilla ESM3 and ESM3+PredMem on a matched length-100 case: whole-chain overlay, backbone detail, and per-residue C$\alpha$ displacement against the ESMFold refolding.}
\label{fig:esm3_case}
\end{figure}

\paragraph{PredMem improves unconditional co-generation with and without guidance.}
\Cref{tab:main} shows that PredMem raises mean success on all three generators,
both without and with CFG. The effect is largest without guidance, where PredMem
lifts success from 33.4\% to 82.1\% on ESM3 and from 28.9\% to 46.6\% on
DPLM-2.1, alongside a smaller +3.6~pp gain on DPLM-2. The gains
are not limited to the thresholded success rate: on ESM3, carrying soft
predictions across steps suppresses the repetitive mode collapse of unguided
vanilla decoding---which otherwise generates low-complexity sequences and
extended, poorly folded backbones---reducing mean scRMSD over all outputs
from 31.9 to 2.3~\AA{}, raising mean scTM from 0.545 to 0.944, and lifting
refolded pLDDT from 67.9 to 90.5. As illustrated for a matched $L=100$
trajectory in \cref{fig:esm3_case}, vanilla ESM3 collapses into a near-homopolymeric
poly-Ala sequence (2 residue types) whose generated rod diverges sharply from
its ESMFold refolding ($\mathrm{scRMSD}=34.1$~\AA{}, $\mathrm{TM}=0.19$),
whereas PredMem yields a compositionally diverse sequence (18 residue types) with
tight sequence--structure agreement ($\mathrm{scRMSD}=0.38$~\AA{},
$\mathrm{TM}=0.99$, $\mathrm{pLDDT}=85.4$).
DPLM-2.1 follows the same trend, with mean scRMSD falling from 5.2 to 3.3~\AA{}. PredMem also remains beneficial on top of CFG, improving success by 2.3 to
16.5 percentage points across models. On ESM3 and DPLM-2.1, PredMem alone already
exceeds CFG, by 17.6 and 13.7 points, while evaluating one branch per step
instead of three. Under the inherited CFG settings, carrying predictions
across steps therefore appears more effective than reshaping logits within
a step on these two generators.

\paragraph{Higher success need not broaden structural coverage.}
The number of successful clusters does not follow success. Without guidance,
PredMem adds clusters on ESM3 and DPLM-2 but loses 17.0 on DPLM-2.1. With CFG, it
loses about 30 on both ESM3 and DPLM-2.1 and changes DPLM-2 by only 1.2. These
results separate improved self-consistency from broader coverage of successful
structures. Section~\ref{sec:analysis} examines this trade-off with additional
controls.

\subsection{Motif Scaffolding}
\label{sec:motif}

\paragraph{Protocol.}
The motif fixes residue identities and structural conditions while the
remaining positions are generated. Each run produces 2,400 scaffolds across
24 motif variants, with 100 samples per variant. Success requires
$\mathrm{scRMSD}<2$~\AA{} and generated-motif $\mathrm{RMSD}<1$~\AA{}
on the decoded backbone, following \citet{zhou2026unlocking}. A variant is solved if any of its
100 scaffolds succeeds. \Cref{tab:motif} reports success, self-consistency,
solved variants, and diversity.

\begin{table*}[t]
\centering
\small
\setlength{\tabcolsep}{4.5pt}
\renewcommand{\arraystretch}{1.06}
\caption{
\textbf{Motif scaffolding.} Each run produces 2,400 scaffolds and success rates are means over five seeds. PredMem raises mean success on all three generators, most strongly on ESM3.}
\label{tab:motif}
\begin{tabular}{@{}ll*{5}{r@{\hspace{2pt}}l}@{}}
\toprule
& & \multicolumn{6}{c}{Quality} & \multicolumn{2}{c}{Coverage} & \multicolumn{2}{c}{Diversity} \\
\cmidrule(lr){3-8} \cmidrule(lr){9-10} \cmidrule(l){11-12}
Model & Method
& \multicolumn{2}{c}{Success (\%) $\uparrow$}
& \multicolumn{2}{c}{scTM $\uparrow$}
& \multicolumn{2}{c}{scRMSD $\downarrow$}
& \multicolumn{2}{c}{Solved\,/\,24 $\uparrow$}
& \multicolumn{2}{c}{\#Clusters $\uparrow$} \\
\midrule
\multirow{4}{*}{DPLM-2}
 & Vanilla & 20.60 &  & 0.666 &  & \best{4.95} &  & \second{16.8} &  & 37.8 & \\
 & PredMem & \best{21.71} & \dPos{+1.11} & \second{0.668} & \dPos{+0.002} & \best{4.95} & \dZero{+0.00} & \best{17.0} & \dPos{+0.2} & \best{41.0} & \dPos{+3.2} \\
 & CFG & 20.48 &  & 0.661 &  & \second{5.03} &  & 16.2 &  & 37.6 & \\
 & CFG+PredMem & \second{21.25} & \dPos{+0.77} & \best{0.669} & \dPos{+0.008} & \best{4.95} & \dPos{-0.08} & 16.4 & \dPos{+0.2} & \second{38.2} & \dPos{+0.6} \\
\midrule
\multirow{4}{*}{DPLM-2.1}
 & Vanilla & 18.33 &  & 0.659 &  & 5.38 &  & 15.2 &  & 30.2 & \\
 & PredMem & \second{20.29} & \dPos{+1.96} & \best{0.671} & \dPos{+0.012} & \best{5.16} & \dPos{-0.22} & \second{15.4} & \dPos{+0.2} & \second{31.2} & \dPos{+1.0} \\
 & CFG & 19.07 &  & 0.652 &  & 5.48 &  & 14.8 &  & 30.0 & \\
 & CFG+PredMem & \best{21.08} & \dPos{+2.01} & \second{0.664} & \dPos{+0.012} & \second{5.23} & \dPos{-0.25} & \best{16.4} & \dPos{+1.6} & \best{36.8} & \dPos{+6.8} \\
\midrule
\multirow{4}{*}{ESM3}
 & Vanilla & 30.43 &  & 0.711 &  & 6.97 &  & 19.4 &  & 57.4 & \\
 & PredMem & \best{42.40} & \dPos{+11.97} & \second{0.767} & \dPos{+0.056} & 4.33 & \dPos{-2.64} & 21.8 & \dPos{+2.4} & 64.8 & \dPos{+7.4} \\
 & CFG & 35.81 &  & 0.754 &  & \second{4.29} &  & \best{22.6} &  & \second{83.6} & \\
 & CFG+PredMem & \second{41.93} & \dPos{+6.12} & \best{0.776} & \dPos{+0.022} & \best{3.75} & \dPos{-0.54} & \second{22.0} & \dNeg{-0.6} & \best{95.0} & \dPos{+11.4} \\
\bottomrule
\end{tabular}
\end{table*}

\paragraph{PredMem improves motif scaffolding across all three generators.}
Adding PredMem increases mean success on every model in \cref{tab:motif}, both
without guidance and on top of CFG. In motif scaffolding,
ESM3 benefits most---rising by 12.0 percentage points without guidance, from
30.4\% to 42.4\%, and cutting mean scRMSD from 7.0 to 4.3~\AA{}---while
DPLM-2 and DPLM-2.1 show smaller gains of 0.8 to 2.0 points. Across all
three generators, unguided PredMem alone also surpasses CFG in mean success.
Crucially, unlike the concentration observed in Section~\ref{sec:cogen}, PredMem expands
successful-cluster counts in all six model--guidance comparisons---adding
11.4 clusters on top of CFG for ESM3 and 6.8 for DPLM-2.1---and solves more
motif variants in five of the six settings. Under motif conditioning,
retaining uncommitted predictions therefore improves both design yield and structural breadth.

\section{Analysis}
\label{sec:analysis}

\Cref{sec:results} shows that PredMem improves co-generation with and without CFG.
We now examine how its feedback changes decoding, whether simpler alternatives 
reproduce the gain, and how it trades off quality against diversity.

\subsection{Early Belief Consolidation}
\label{sec:early_dynamics}

\begin{figure*}[t]
    \centering
    \begin{minipage}[t]{\columnwidth}
        \centering
        \includegraphics[width=\linewidth]{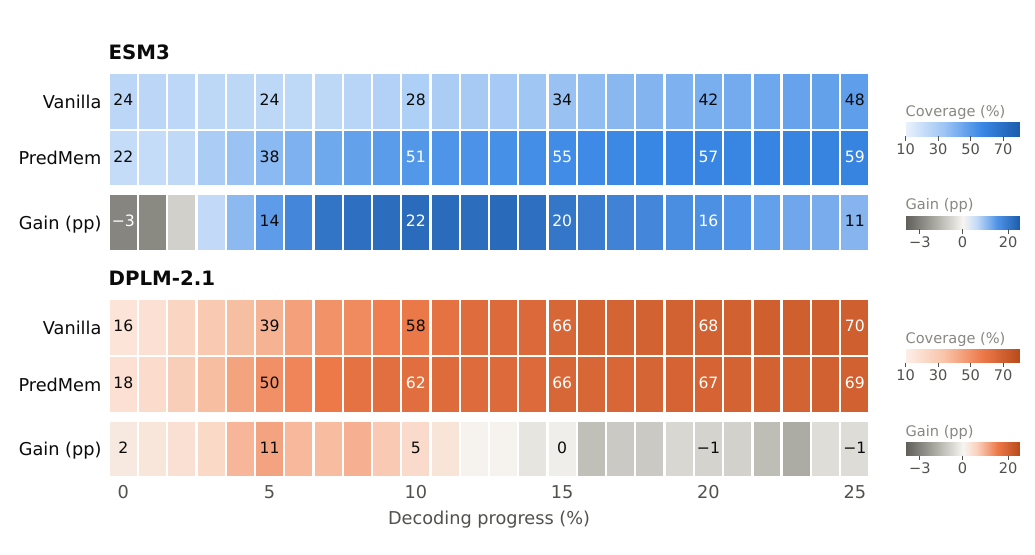}
        \caption{\textbf{PredMem accelerates early consensus toward terminal tokens.}
Top-3 coverage over the first 25\% of decoding. Gain rows report PredMem minus vanilla in percentage points.}
        \label{fig:early}
    \end{minipage}\hfill
    \begin{minipage}[t]{\columnwidth}
        \centering
        \includegraphics[width=0.82\linewidth]{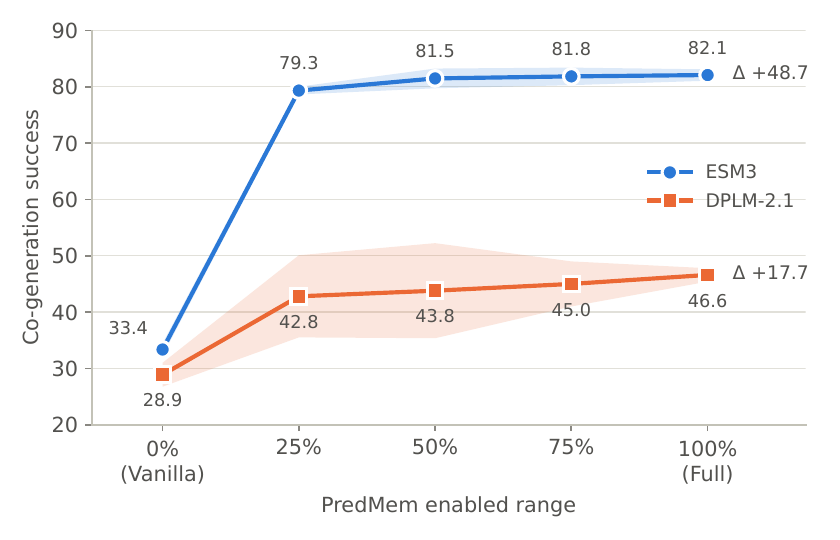}
        \caption{\textbf{Functional contribution across decoding windows.}
 Final co-generation success versus the initial fraction of steps with PredMem enabled.}
        \label{fig:pm_windows_unguided}
    \end{minipage}
\end{figure*}

\paragraph{PredMem front-loads trajectory convergence toward terminal tokens.}
Rather than acting as a late-stage refiner, PredMem functions as an early belief consolidator that steers unresolved positions toward self-consistent final states while committed context is still sparse. To observe this mechanism directly, we track \emph{Top-3 coverage} in \cref{fig:early}---the fraction of still-masked positions whose eventual final token already appears among the current Top-3 predictions. Across both backbones, PredMem accelerates this internal consensus early in decoding, though with distinct dynamics. On ESM3, vanilla decoding stalls at low coverage during the opening phase, whereas PredMem breaks this plateau within the first $10\%$ of steps, boosting coverage by up to $22$  percentage points and sustaining a double-digit lead throughout the first quarter. On DPLM-2.1, where vanilla decoding converges faster on its own, PredMem jump-starts consensus in the earliest steps---lifting coverage by $11$ percentage points at $5\%$ progress---before vanilla catches up.

\paragraph{Bootstrapping early consensus explains most of the quality gain.}
Because PredMem mainly stabilizes cross-modal hypotheses before many hard tokens are committed, its impact is concentrated in the opening phase of decoding. As shown in \cref{fig:pm_windows_unguided}, enabling PredMem only during the first $25\%$ of steps and then completing the trajectory under vanilla decoding already recovers roughly $94\%$ of the full-run success gain on ESM3 and $79\%$ on DPLM-2.1. Once enough hard tokens anchor the sequence--structure context, the marginal value of soft feedback diminishes, and the run-to-run uncertainty bands overlap substantially when extending PredMem beyond the first $25\%$. This also accounts for the smaller gains in motif scaffolding (\cref{sec:motif}), where the fixed motif already supplies committed cross-modal anchors from the first step.

\subsection{Ruling Out Alternative Explanations}
\label{sec:analysis_alternatives}

We test whether extra denoising steps, non-specific memory feedback, or
temperature scaling reproduce PredMem's gains under unguided decoding
(\cref{tab:counterfactual_controls}, using the five-seed budget of
Section~\ref{sec:cogen}). Wall-clock decoding time is measured on a single A800 GPU
at batch size $1$ across all five lengths ($10$ repeats each, excluding model
loading and refolding).

\begin{table}[t]
\centering
\small
\setlength{\tabcolsep}{3pt}
\renewcommand{\arraystretch}{1.06}
\caption{\textbf{Controls under unguided decoding: success and cost.}
 Five-seed mean success (Section~\ref{sec:cogen} protocol) and decoding time; dashes mark unmeasured entries. Standard deviations, deltas, and control definitions: Appendix~\ref{app:analysis_controls}.}
\label{tab:counterfactual_controls}
\footnotesize
\begin{tabular}{@{}l cc @{\hspace{8pt}} cc@{}}
\toprule
& \multicolumn{2}{c}{ESM3} & \multicolumn{2}{c}{DPLM-2.1} \\
\cmidrule(lr){2-3} \cmidrule(lr){4-5}
Method
& Success (\%) $\uparrow$ & Time (s) $\downarrow$
& Success (\%) $\uparrow$ & Time (s) $\downarrow$ \\
\midrule
Vanilla & 33.4 & 17.86 & 28.9 & 24.83 \\
Vanilla~$2T$ & 34.5 & 35.74 & 30.4 & 49.59 \\
PredMem+shift & 48.8 & --- & 35.1 & --- \\
PredMem+floor & 72.5 & --- & 36.0 & --- \\
\textbf{PredMem} & \textbf{82.1} & 18.42 & \textbf{46.6} & 28.02 \\
\bottomrule
\end{tabular}
\end{table}

\paragraph{Extra decoding steps do not reproduce the gain.}
Doubling the step budget without memory in Vanilla~$2T$ doubles decoding
time yet yields negligible success changes within seed variability on both
backbones (\cref{tab:counterfactual_controls}). By contrast, PredMem achieves its
full gains with only $3.1\%$ (ESM3) and $12.8\%$ (DPLM-2.1) extra decoding
time and at most $0.12$~GiB of peak memory overhead, ruling out additional
compute as the driver.

\paragraph{Positional correspondence matters on both backbones.}
Cyclically shifting memory across unresolved sites in PredMem+shift preserves
the multiset of cached summaries while breaking residue-level alignment.
This eliminates roughly two-thirds of PredMem's gain on both backbones,
leaving only a smaller residual gain from global regularization.

\paragraph{PredMem reshapes token rankings beyond entropy and temperature scaling.}
To test whether distribution sharpening alone drives the gain, PredMem+floor
raises temperature to match the no-memory reference entropy whenever PredMem's
entropy drops below it, preserving token rankings. Both backbones still
substantially outperform Vanilla, retaining roughly $80\%$ (ESM3) and $40\%$
(DPLM-2.1) of PredMem's gain. Consistently, temperature sweeps show that while
retuning Vanilla recovers part of DPLM-2.1's default-temperature gain, PredMem
maintains a clear lead across temperatures (\cref{sec:analysis_diversity}). Moreover, paired logits at
identical decoding states reveal rank changes impossible under positive
scalar temperature rescaling: at roughly $25\%$ progress, PredMem alters ESM3's
Top-1 prediction at $17.3\%$ of sequence and $62.9\%$ of structure positions. Together, these controls establish
that while entropy reduction contributes partially on DPLM-2.1, PredMem's primary
mechanism is position-aligned belief refinement that reshapes token rankings (details in Appendix~\ref{app:analysis_controls}).

\begin{figure}[t]
\centering
\includegraphics[width=\columnwidth]{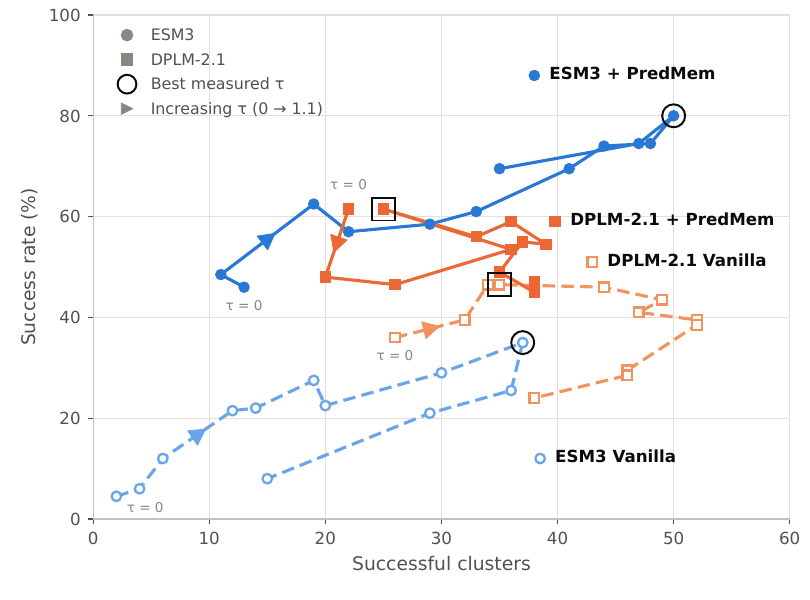}
\caption{\textbf{Success–diversity Pareto frontiers across decoding temperatures.}
Unguided decoding sweeps over 12 temperatures $\tau \in [0, 1.1]$ for vanilla decoding and PredMem; circles denote ESM3, squares DPLM-2.1.}
\label{fig:frontier_tau}
\end{figure}

\subsection{Quality, Diversity, and the Pareto Frontier}
\label{sec:analysis_diversity}
\label{sec:analysis_tradeoff}

Generative protein design faces an inherent trade-off between structural
viability and output diversity. By steering early decoding steps toward
self-consistent states, PredMem channels trajectories into high-viability fold
basins, substantially elevating design success while maintaining strong
structural coverage.

\paragraph{Pushing the Pareto frontier of success and structural diversity.}
To examine this trade-off, we sweep the constant sampling temperature $\tau$
across $12$ operating points from $0.0$ to $1.1$ under unguided decoding,
evaluating $200$ candidates per setting via Foldseek clustering at a
TM-score threshold of $0.5$ (\cref{fig:frontier_tau}). Across both
backbones, PredMem shifts the quality--diversity frontier markedly upward:
\begin{itemize}
    \item \textbf{Strict Pareto dominance on ESM3.} The \texttt{ESM3+PredMem}
    trajectory lies strictly above and to the right of \texttt{ESM3 Vanilla}
    across all temperatures. Rather than trading diversity for quality, PredMem
    improves both simultaneously: at their respective peak-success temperatures,
    PredMem more than doubles the success rate from $35.0\%$ to $80.0\%$ while
    expanding successful cluster yields from $37$ to $50$.
    \item \textbf{High-success regime shift on DPLM-2.1.} On DPLM-2.1, PredMem
    lifts success above $60\%$, compared with at most $46.5\%$ for vanilla
    decoding. Although earlier basin convergence caps the maximum cluster
    count at high temperatures, \texttt{DPLM-2.1+PredMem} achieves a substantially
    higher success rate at every overlapping diversity level, offering a more
    productive operating regime when designability is the primary bottleneck.
\end{itemize}

\section{Conclusion}
\label{sec:conclusion}

We introduced Predictive Memory (PredMem), a training-free inference-time
recurrence mechanism that addresses the memoryless mask reset in iterative
masked multimodal protein generation. By carrying probability-weighted Top-$k$
expectation embeddings across decoding steps with strict branch-local
isolation under classifier-free guidance, PredMem substantially boosts
sequence--structure self-consistency and design success across ESM3, DPLM-2,
and DPLM-2.1 with negligible compute and memory overhead. Beyond driving strong
gains in unconditional co-generation and ESM3 motif scaffolding, PredMem
suppresses repetitive collapse and pushes the quality--diversity Pareto
frontier upward. While our study uses a fixed recurrence schedule on
frozen models and relies on \emph{in silico} self-consistency benchmarks,
future work will explore uncertainty-adaptive gating and extend this recurrent
soft-belief mechanism to experimental functional design, such as
target-specific \emph{de novo} antibody sequence--structure co-design.

\section*{LLM Usage Statement}

We used LLM tools in two limited ways. First, GPT and Codex helped polish the English writing for clarity and grammar. Second, Codex helped us debug our code. Beyond these uses, all research content is the authors' own work. This includes the idea of Predictive Memory, the mathematical derivations, and the inference hooks for ESM3, DPLM-2, and DPLM-2.1. It also includes the experimental design and the interpretation of all results. To ensure accuracy, we reviewed every LLM-suggested edit to the text and code, and verified these edits against the raw experimental outputs and the original references. Therefore, the authors take full responsibility for the entire manuscript.
{
    \small
    \bibliographystyle{ieeenat_fullname}
    \bibliography{predictive}
}

\clearpage
\appendix
\section{Implementation Details: Branch Inputs and Read Interfaces}
\label{app:pm_interfaces}

\subsection{Branch Inputs and Protected Positions}
\label{app:pm_branches}

Branch construction is a task-specific map
$(z_t^b,c^b)=\mathcal D_b(z_t,c)$ shared by each decoder and
its PredMem-augmented counterpart.
The reference unconditional co-generation construction is
\begin{equation}
  z_t^+=(s_t,x_t),\qquad
  z_t^{u_S}=(s_t,\mathrm{null}),\qquad
  z_t^{u_X}=(\mathrm{null},x_t),
  \label{eq:pm_branch_inputs}
\end{equation}
where $\mathrm{null}$ denotes the native dropped-condition representation
rather than removal of residue positions~\cite{zhou2026unlocking}.
For each branch, only the pairs in Eq.~\eqref{eq:pm_pairs} receive
feedback. In single-target tasks, only the corresponding two pairs are
used, or only $(+,m)$ when CFG is disabled.

Generation eligibility and feedback eligibility are distinct.
For example, coordinate conditioning may protect a motif's input
coordinates while its structure tokens remain generation targets;
such protected positions can belong to $\mathcal I_X^{\mathrm{gen}}$
while remaining outside $\mathcal I_X^{\mathrm{fb}}$.
For coordinates-only motif conditioning, positions with finite
conditioning coordinates, fixed amino acids, and BOS/EOS tokens remain
outside $\mathcal I_X^{\mathrm{fb}}$ and are excluded from both memory
feedback and reads. The input coordinates are preserved, and decoded
motif geometry is evaluated without clamping to the reference.

\subsection{Token-Space Read for ESM3}
\label{app:pm_esm3}

At an eligible masked position, Eq.~\eqref{eq:pm_read} replaces the
native modality-specific token-embedding component while leaving other
input components and subsequent native processing in place.
Visible tokens and protected conditioning inputs remain unchanged.
Top-$k$ selection, softmax normalization, and embedding aggregation are
computed in FP32, and caches are detached and cast to the input dtype at
the read interface.

\subsection{Read Maps for DPLM Variants}
\label{app:pm_dplm}

Suppressing branch, modality, position, and time indices for clarity,
let $h_0$ be the native post-dropout input at an eligible position and
$\mu_\kappa=\kappa\mu$ the cached summary $\mu$ of
Eq.~\eqref{eq:pm_write} after the appropriate native scaling. A
pre-LayerNorm read forms
\begin{equation}
  \widetilde h=(1-\lambda)h_0+\lambda\mu_\kappa.
  \label{eq:pm_preln}
\end{equation}
The first attention block receives $\operatorname{LN}(\widetilde h)$,
and the residual stream uses $\widetilde h$.
A post-LayerNorm read instead forms
\begin{equation}
  \widetilde u
  =(1-\lambda)\operatorname{LN}(h_0)
   +\lambda\operatorname{LN}(\mu_\kappa),
  \label{eq:pm_postln}
\end{equation}
which replaces the normalized attention input while retaining $h_0$
on the residual path. A combined read uses
Eq.~\eqref{eq:pm_postln} for attention and
Eq.~\eqref{eq:pm_preln} for the residual stream.
These are distinct read maps; the DPLM experiments use the
post-LayerNorm read of Eq.~\eqref{eq:pm_postln}
(Appendix~\ref{app:exp_memory}).

\paragraph{Native scaling.}
When the native token-dropout path uses mask-ratio rescaling, write
\begin{equation}
  s(I)=\frac{1-\rho_{\mathrm{train}}}
  {1-n_{\mathrm{mask}}(I)/n_{\mathrm{nonpad}}(I)},
  \label{eq:pm_scale}
\end{equation}
on inputs for which the denominator is positive, where
$\rho_{\mathrm{train}}=0.12$ is the native training mask-rate constant.
Let $s_2$ be the scale evaluated on the concatenated input and $s_1$ the
scale evaluated on the sequence half. The DPLM-2 embedding path gives
$\kappa_S=\kappa_X=s_2^2$, whereas for DPLM-2.1, $\kappa_S=s_1s_2$ and
$\kappa_X=s_2$. When token dropout is disabled, these scaling factors are
one. The multiplier $\kappa$ combines the scaling stages between the
cached token embedding and the selected read interface, evaluated on the
current branch input at read time.

\paragraph{Effect of normalization.}
For LayerNorm gain $\gamma$, bias $\beta_{\mathrm{LN}}$, and
$\epsilon_{\mathrm{LN}}>0$, let $\bar\mu_\kappa$ and
$\operatorname{Var}(\mu_\kappa)$ be the mean and variance of the entries
of $\mu_\kappa$, and consider a position with $h_0=0$.
Under the pre-LayerNorm read of Eq.~\eqref{eq:pm_preln}, the attention
input satisfies
\begin{equation}
  \operatorname{LN}(\lambda\mu_\kappa)-\operatorname{LN}(0)
   =\gamma\odot
  \frac{\lambda(\mu_\kappa-\bar\mu_\kappa\mathbf 1)}
  {\sqrt{\lambda^2\operatorname{Var}(\mu_\kappa)+\epsilon_{\mathrm{LN}}}}.
  \label{eq:pm_ln_effect}
\end{equation}
The affine bias cancels, and for positive $\lambda$ with
$\lambda^2\operatorname{Var}(\mu_\kappa)\gg\epsilon_{\mathrm{LN}}$,
normalization approximately cancels the nominal amplitude scaling in the
attention input. Under the post-LayerNorm read of
Eq.~\eqref{eq:pm_postln}, the same position gives
\begin{equation}
  \widetilde u-\operatorname{LN}(0)
   =\lambda\,\gamma\odot
  \frac{\mu_\kappa-\bar\mu_\kappa\mathbf 1}
  {\sqrt{\operatorname{Var}(\mu_\kappa)+\epsilon_{\mathrm{LN}}}},
  \label{eq:pm_postln_effect}
\end{equation}
which scales linearly in $\lambda$ and is not cancelled by
normalization. The two reads therefore respond differently to $\lambda$
at the attention input, which explains why the read interface must be
explicitly specified and motivates the ablation in
Appendix~\ref{app:exp_memory}.

\subsection{Complete-Code Memory for DPLM-2.1}
\label{app:pm_codes}

DPLM-2.1 represents a structure token using 13 binary components.
Let $d_j(v)\in\{0,1\}$ be the label of component $j$ for complete
code $v$, and let $p_{t,i,j}^{b,X}(a)$ be the unguided probability
of label $a$ from that component's output head.
Over all $2^{13}$ complete codes, define
\begin{equation}
  \ell_{t,i}^{b,X}(v)
  :=\log q_{t,i}^{b,X}(v)
  =\sum_{j=1}^{13}
  \log p_{t,i,j}^{b,X}\!\left(d_j(v)\right).
  \label{eq:pm_code_scores}
\end{equation}
This is the factorized distribution specified by the output heads.
We select top-$k$ complete codes under these scores and apply
Eqs.~\eqref{eq:pm_topk} and~\eqref{eq:pm_write} using
\begin{equation}
  E_X(v)=\operatorname{quant2emb}(\mathbf d(v)),
  \label{eq:pm_code_embedding}
\end{equation}
where $\mathbf d(v)=(d_1(v),\ldots,d_{13}(v))$ is the native input
representation of the bit pattern, including its sign convention and
scaling, with the projection bias retained. Complete-code scores are
evaluated in FP32 with autocast disabled, using the native sampler's
code-to-bit ordering. Memory construction uses probabilities before CFG
and at cache temperature one, while retaining the native hard-token
sampler.

\section{Properties of Predictive Memory}
\label{app:pm_analysis}

The following analysis characterizes the approximation fidelity,
variance-reduction behavior, and distributional stability of predictive
memory at the affine token-space read interface.

\subsection{\texorpdfstring{Top-$k$}{Top-k} Truncation}
\label{app:pm_truncation}

Suppress the branch, modality, position, and time indices for clarity.
Let $p=\operatorname{softmax}(\ell)$ be the unit-temperature distribution
over $\mathcal V_m$ that underlies Eqs.~\eqref{eq:pm_topk}
and~\eqref{eq:pm_write}. Let $K$ be the top-$k$ set and
$1-\varepsilon_K=\sum_{v\in K}p(v)>0$ its retained mass, so that
$\widetilde p(v)=p(v)/(1-\varepsilon_K)$ on $K$.
Define the full-vocabulary mean and the cached summary
\begin{equation}
  \mu^{\mathrm{full}}=\sum_{v\in\mathcal V_m}p(v)E_m(v),
  \qquad
  \mu=\sum_{v\in K}\frac{p(v)}{1-\varepsilon_K}E_m(v).
  \label{eq:pm_full_and_topk}
\end{equation}
For embedding diameter
$D_m=\max_{u,v\in\mathcal V_m}\|E_m(u)-E_m(v)\|_2$,
\begin{equation}
  \|\mu-\mu^{\mathrm{full}}\|_2\le\varepsilon_K D_m.
  \label{eq:pm_truncation_bound}
\end{equation}
\emph{Proof.}
If $\varepsilon_K>0$, let $\mu_{\mathrm{tail}}$ be the conditional
mean outside $K$. Then
$\mu^{\mathrm{full}}=(1-\varepsilon_K)\mu+\varepsilon_K\mu_{\mathrm{tail}}$.
Both conditional means lie in the embedding convex hull, whose diameter
is at most $D_m$. The zero-tail case is immediate.

For the affine read in Eq.~\eqref{eq:pm_read}, the difference between
top-$k$ and full-mean inputs is therefore at most
$\lambda\varepsilon_K D_m$.
This bound isolates the discarded tail probability mass
$\varepsilon_K$---rather than the absolute vocabulary size---as the
quantity governing truncation error, explaining why a fixed small $k$
remains faithful whenever the predictive distribution is concentrated.

\subsection{Mean and Sampled Predictive Inputs}
\label{app:pm_sampled}

Fix the hard state, cache distributions, and eligible positions.
Index the active triples $(b,m,i)$, with $(b,m)\in\mathcal B$ and
$i\in\mathcal A_t^{b,m}$, by $j$, and write
$\widetilde p_j=\widetilde p_{t-1,i}^{b,m}$,
$\mu_j=\mu_{t-1,i}^{b,m}$, and $E_j=E_m$.
Let $V_j\sim\widetilde p_j$, so that
$\mu_j=\sum_v\widetilde p_j(v)E_j(v)$ as in Eq.~\eqref{eq:pm_write}.
For the affine token-space read of Eq.~\eqref{eq:pm_read}, define
\begin{equation}
  \begin{aligned}
    u_j^{\mathrm{draw}} &=(1-\lambda)E_j(\mathrm{MASK})+\lambda E_j(V_j),\\
    \bar u_j &=(1-\lambda)E_j(\mathrm{MASK})+\lambda\mu_j.
  \end{aligned}
  \label{eq:pm_sampled_input}
\end{equation}
Here $\bar u_j=\widetilde e_{t,i}^{b,m}$ is the read of
Eq.~\eqref{eq:pm_read}. Let $U^{\mathrm{draw}}$ and $\bar U$ concatenate
these components, with all remaining inputs held fixed. Their first and
second moments are
\begin{equation}
  \begin{aligned}
    \mathbb E U^{\mathrm{draw}}&=\bar U,\\
    \mathcal V_{\mathrm{draw}}
    &:=\mathbb E\|U^{\mathrm{draw}}-\bar U\|_2^2\\
    &=\lambda^2\sum_j\left[
        \sum_v\widetilde p_j(v)\|E_j(v)\|_2^2
       -\|\mu_j\|_2^2\right].
  \end{aligned}
  \label{eq:pm_sampled_variance}
\end{equation}
The variance identity uses only marginal moments and does not require
independent draws across branches.

For a scalar network output $F$, suppose
$\|\nabla^2F(U)\|_{\mathrm{op}}\le H_F$ along every segment from
$\bar U$ to a possible $U^{\mathrm{draw}}$. Then
\begin{equation}
  \left|\mathbb E F(U^{\mathrm{draw}})-F(\bar U)\right|
  \le\frac{H_F}{2}\mathcal V_{\mathrm{draw}}.
  \label{eq:pm_mean_taylor}
\end{equation}
\emph{Proof.}
The linear Taylor term has zero expectation because
$\mathbb E(U^{\mathrm{draw}}-\bar U)=0$.
The absolute remainder is at most
$H_F\|U^{\mathrm{draw}}-\bar U\|_2^2/2$; taking expectations
gives the result.

Equations~\eqref{eq:pm_sampled_variance} and~\eqref{eq:pm_mean_taylor}
show that replacing stochastic draws $U^{\mathrm{draw}}$ with the
deterministic summary $\bar U$ agrees to first order while eliminating
the second-order perturbation bias driven by sampling variance
$\mathcal V_{\mathrm{draw}}$.
Under a nonlinear post-LayerNorm read, averaging normalized embeddings
and normalizing the averaged embedding do not strictly commute:
\begin{equation}
  \mathbb E\operatorname{LN}(\kappa E_j(V_j))
  \ne\operatorname{LN}(\kappa\mu_j)
  \quad\text{in general}.
  \label{eq:pm_nonlinear_mean}
\end{equation}
Nonetheless, aggregating embeddings prior to normalization in
$\operatorname{LN}(\kappa\mu_j)$ acts as directional pooling, analogously
suppressing the discrete directional-jump variance introduced by a single
stochastic draw $V_j$.

\subsection{Same-State Distribution Changes}
\label{app:pm_softmax}

Consider a single categorical output distribution at a fixed history.
Let $p=\operatorname{softmax}(g^0/\tau)$ and
$q=\operatorname{softmax}(g^F/\tau)$ be the no-memory and
memory-enabled distributions over the same finite vocabulary,
with finite logits and $\tau>0$.
Define
\begin{equation}
  \delta_v=\frac{g^F(v)-g^0(v)}{\tau},
  \qquad r=\max_v\delta_v-\min_v\delta_v.
  \label{eq:pm_logit_range}
\end{equation}
The probability ratio and distributional bounds are
\begin{align}
  q(v)&=\frac{p(v)e^{\delta_v}}{\mathbb E_p e^\delta},
  \label{eq:pm_tilt}\\
  \operatorname{TV}(q,p)&\le\tanh(r/4),
  \label{eq:pm_tv}\\
  \max\{\operatorname{KL}(q\|p),\operatorname{KL}(p\|q)\}
  &\le r^2/8.
  \label{eq:pm_kl}
\end{align}
Here $\operatorname{TV}(q,p)=\frac12\sum_v|q(v)-p(v)|$, and
all logarithms are natural.

\emph{Proof.}
Equation~\eqref{eq:pm_tilt} follows by dividing the softmax
probabilities. Put $X=e^{\delta-\min_v\delta_v}\in[1,e^r]$
and $Z=\mathbb E_pX$. For $r>0$, the chord bound for $|X-Z|$
gives
\begin{equation}
  \begin{aligned}
    \operatorname{TV}(q,p)
    &=\frac{\mathbb E_p|X-Z|}{2Z}\\
    &\le\frac{(e^r-Z)(Z-1)}{Z(e^r-1)}
    \le\tanh(r/4).
  \end{aligned}
  \label{eq:pm_tv_proof}
\end{equation}
The middle expression is maximized at $Z=e^{r/2}$.
For KL, let $\phi(s)=\log\mathbb E_p e^{s\delta}$.
Its second derivative is a variance under exponential tilting,
so $\phi''(s)\le r^2/4$.
Specifically,
$\operatorname{KL}(p\|q)=\int_0^1(1-s)\phi''(s)\,\mathrm ds$
and
$\operatorname{KL}(q\|p)=\int_0^1s\phi''(s)\,\mathrm ds$,
yielding Eq.~\eqref{eq:pm_kl}. If $r=0$, then $p=q$.

For any continuation quality metric $Q:\mathcal V\to[0,1]$,
Eq.~\eqref{eq:pm_tilt} yields
\begin{equation}
  \mathbb E_qQ-\mathbb E_pQ
  =\frac{\operatorname{Cov}_p(Q,e^\delta)}
  {\mathbb E_p e^\delta}.
  \label{eq:pm_quality_covariance}
\end{equation}
This covariance identity isolates the exact mechanism through which
memory feedback improves generation: expected quality increases
whenever the induced logit shift $\delta$ correlates positively with
high-quality continuations ($\operatorname{Cov}_p(Q,e^\delta)>0$),
providing a theoretical anchor for confidence-gated and branch-coupled
feedback, while Eqs.~\eqref{eq:pm_tv} and~\eqref{eq:pm_kl} ensure that
the single-step distributional shift remains smoothly bounded by $r$.

\FloatBarrier
\section{Models, Tasks, and Evaluation}
\label{app:models_tasks}
\label{app:exp_metrics}

\subsection{Pretrained Generators}
We evaluate three frozen multimodal protein generators, following the task definitions and decoding terminology of \citet{zhou2026unlocking}. The evaluated configurations (Vanilla, PredMem, CFG, and CFG+PredMem for the joint tasks) and their hyperparameters are detailed in Appendix~\ref{app:exp_protocol}.

\paragraph{DPLM-2.}
DPLM-2 jointly models amino-acid sequences and discrete structure tokens via masked diffusion~\citep{wang2025dplm2}. We use the 650M-parameter checkpoint \texttt{airkingbd/dplm2\_650m} together with its structure tokenizer, \texttt{airkingbd/struct\_tokenizer}. Sequence and structure tokens share a unified denoising trajectory while maintaining modality-specific vocabularies: sequence conditions are supplied as amino-acid tokens, and structural conditions use the native tokenizer. PredMem interpolates at the post-LayerNorm attention input after matching the native scaling factor (Appendix~\ref{app:pm_dplm}).

\paragraph{DPLM-2.1.}
We use the 650M-parameter binary-structure variant \texttt{airkingbd/dplm2\_bit\_650m}~\citep{hsieh2025designspace}, whose structural head represents each token through 13 binary components, yielding $2^{13}=8192$ complete codes. PredMem ranks complete codes under the factorized output distribution and maps each selected code through the native structure-input projection; thus, $k$ denotes complete structural codes rather than independently selected bits (Appendix~\ref{app:pm_codes}). The read interface is the same post-LayerNorm attention read used for DPLM-2.

\paragraph{ESM3.}
We use the open 1.4B-parameter ESM3 checkpoint \texttt{EvolutionaryScale/esm3-sm-open-v1}~\citep{hayes2025simulating}. Although ESM3 supports several aligned protein tracks, our experiments generate sequence and structure, using its native coordinate-conditioning interface for inverse folding and motif scaffolding. PredMem applies the token-space read directly to the sequence and structure token embeddings (Appendix~\ref{app:pm_esm3}), leaving all other input components and frozen generator parameters unchanged.

\subsection{Tasks and Datasets}
\paragraph{Unconditional co-generation.}
Both tracks start fully masked without external sequence or structure conditioning. Each run generates 500 sequence--structure pairs per configuration (100 at each length in $\{100,200,300,400,500\}$), updating both tracks synchronously in the main experiments.

\paragraph{Motif scaffolding.}
We evaluate 24 motif variants with 100 scaffolds per variant per run, totaling 2,400 scaffolds per configuration. The motif specifies fixed residue identities and structural constraints while the remaining positions are generated. DPLM variants receive tokenized motif conditions, whereas ESM3 uses coordinate conditioning. Fixed sequence positions and supplied coordinates are preserved throughout decoding, and decoded motif geometry is evaluated directly without clamping to the reference coordinates.

\paragraph{Structure prediction.}
This single-modality conditional task fixes the amino-acid sequence and generates the structure. We evaluate on the full CAMEO2022 set (183 targets) and the PDB Date Split set (449 targets), comparing predicted structures directly against reference structures.

\paragraph{Inverse folding.}
This task~\citep{hsu2022inverse} fixes the backbone structure and generates the sequence on the same two target sets. We measure amino-acid recovery and independently refold each generated sequence to assess consistency with the conditioning structure. CFG is disabled in both single-modality conditional tasks.

\subsection{Metrics}
\paragraph{Structural quality and self-consistency.}
ESMFold~\citep{lin2023evolutionary} independently refolds all generated sequences. For joint tasks, scTM (TM-score~\citep{zhang2004tmscore}) and scRMSD compare the refolded structure against the generated backbone; for inverse folding, they compare it against the conditioning structure. Structure-prediction TM-score and RMSD instead compare predicted and reference structures. Quality metrics are averaged over all generated outputs. For unconditional co-generation and motif scaffolding, pLDDT denotes the mean per-residue confidence of the independent ESMFold refolding output on a 0--100 scale (rather than internal confidence from the generator's structure decoder). TM-score is computed over the complete C$\alpha$ trace, whereas RMSD follows each benchmark's backbone mask and alignment convention.

\paragraph{Success and motif coverage.}
Unconditional co-generation succeeds when $\mathrm{scRMSD}<2$~\AA, whereas motif scaffolding additionally requires generated-motif $\mathrm{RMSD}<1$~\AA. Let $x^{\rm gen}$ denote the generator-decoded coordinates, $x^{\rm EF}$ the ESMFold prediction of the generated sequence, $x^{\rm ref}$ the reference motif, and $\mathcal I^{\rm motif}$ the motif residue indices. The generated-motif RMSD is
\begin{equation}
  R_{\rm gen} = \operatorname{RMSD}(x^{\rm gen}_{\mathcal I^{\rm motif}},x^{\rm ref}),
\end{equation}
computed over aligned backbone atoms after concatenating discontiguous motif segments. The scaffolding success indicator is thus
\begin{equation}
  \mathbf{1}[\mathrm{scRMSD}(x^{\rm gen},x^{\rm EF})<2\,\text{\AA}]\,
  \mathbf{1}[R_{\rm gen}<1\,\text{\AA}].
\end{equation}
Refolded pLDDT is reported alongside $R_{\rm gen}$ as a complementary quality diagnostic. A motif variant is counted as solved if at least one of its 100 scaffolds succeeds, yielding a run-level count of at most 24.

\paragraph{Structural diversity.}
Foldseek~\citep{van2024fast} clustering at a TM-score threshold of 0.5 is computed strictly over the successful subset. We cluster within each length for unconditional co-generation and within each motif variant for scaffolding, then sum cluster counts across a run; this total is reported as \#Clusters. All runs use Foldseek alignment type 1, coverage mode 0, and minimum sequence identity 0.

\paragraph{Sequence recovery.}
For a target of length $L$, with $s_i$ and $s_i^{\rm ref}$ denoting the generated and reference amino acids at position $i$, amino-acid recovery is $100L^{-1}\sum_{i=1}^{L}\mathbf{1}[s_i=s_i^{\rm ref}]$. We report the unweighted (macro) average of target-level recovery across each dataset.

\paragraph{Aggregation and variation.}
For run-level values $y_1,\ldots,y_5$ across five seeds, we report
\begin{equation}
 \bar y=\frac{1}{5}\sum_{r=1}^{5}y_r,\qquad
 s_y=\sqrt{\frac{1}{4}\sum_{r=1}^{5}(y_r-\bar y)^2}.
\end{equation}
Success rates, solved variants, and cluster counts are computed within each run prior to aggregation. The $\pm$ values in Appendix~\ref{app:full_results} denote run-level sample standard deviations (rather than standard errors or per-protein variation); deterministic decoding configurations naturally yield zero run-level variation. Main-table entries are reported at their displayed rounding precision.

\FloatBarrier
\section{Decoding and Hyperparameter Settings}
\label{app:exp_protocol}
\label{app:exp_sampling}
Settings are organized by task using the model and method names in Tables~\ref{tab:main}, \ref{tab:motif}, and~\ref{tab:full_conditional}. \emph{Default} denotes the standard base configuration, whereas \emph{Vanilla} denotes the configured decoder without memory or guidance. PredMem disables guidance, CFG disables memory, and CFG+PredMem enables both; N/A marks parameters that are inactive when memory is disabled.

Let $L$ denote the total sequence length. The support $k$ is the number of predictions retained in memory (rather than token-sampling top-$k$), $\lambda$ is the memory interpolation weight, and $(w_S,w_X)$ are the sequence and structure CFG strengths. Disabled CFG uses a single conditional branch and the conditioned logits directly, corresponding to $w_m=1$ in Eq.~\eqref{eq:pm_cfg} rather than setting $w_m=0$.

\subsection{Unconditional Co-generation}
\label{app:config_joint}
Table~\ref{tab:config_joint} specifies the four arms of Table~\ref{tab:main}. Both modalities are decoded synchronously with remasking. Within each model, all arms share the exact same step budget, position-selection rule, and sampling-temperature schedule.
\begin{table*}[t]
\centering\small
\setlength{\tabcolsep}{5pt}
\renewcommand{\arraystretch}{1.08}
\begin{tabular}{llccc}
\toprule
Setting & Arm / dataset & DPLM-2 & DPLM-2.1 & ESM3\\
\midrule
Order & All & synchronous & synchronous & synchronous\\
Steps & All & 500 & 500 & $L$\\
Unmasking & All & stochastic & stochastic & stochastic\\
Remasking & All & enabled & enabled & enabled\\
Temperature parameter & All & $2.0\to0.1$ & $1.1\to0.1$ & $1.0\to0$\\
\midrule
Memory $(\lambda,k)$ & Vanilla / CFG & $(0,\mathrm{N/A})$ & $(0,\mathrm{N/A})$ & $(0,\mathrm{N/A})$\\
Memory $(\lambda,k)$ & PredMem & $(0.03,3)$ & $(0.3,3)$ & $(0.3,3)$\\
Memory $(\lambda,k)$ & CFG+PredMem & $(0.03,3)$ & $(0.3,3)$ & $(0.3,3)$\\
CFG $(w_S,w_X)$ & Vanilla / PredMem & disabled & disabled & disabled\\
CFG $(w_S,w_X)$ & CFG / CFG+PredMem & $(2.0,1.5)$ & $(2.0,1.5)$ & $(2.0,3.0)$\\
Read interface & PredMem / CFG+PredMem & post-LN attention & post-LN attention & token embedding\\
\bottomrule
\end{tabular}
\caption{\textbf{Unconditional co-generation settings.} Decoding, PredMem, and CFG hyperparameters for each model. Temperature entries list sampler parameters; the resulting schedules are defined in Section~\ref{app:temperature}.}
\label{tab:config_joint}
\end{table*}

\subsection{Motif Scaffolding}
\label{app:config_motif}
Table~\ref{tab:config_motif} specifies the settings for Table~\ref{tab:motif}. To ensure a unified comparison across all three generators and four arms, our main experiments use synchronous decoding throughout (whereas the reference Default recipe in \citet{zhou2026unlocking} uses sequential ESM3 decoding). DPLM-2 and ESM3 set their step budgets to the total sequence length $L$.
\begin{table*}[t]
\centering\small
\setlength{\tabcolsep}{5pt}
\renewcommand{\arraystretch}{1.08}
\begin{tabular}{llccc}
\toprule
Setting & Arm / dataset & DPLM-2 & DPLM-2.1 & ESM3\\
\midrule
Order & All & synchronous & synchronous & synchronous\\
Steps & All & $L$ & 500 & $L$\\
Unmasking & All & stochastic & stochastic & random\\
Remasking & All & enabled & enabled & enabled\\
Temperature parameter & All & $0.5\to0$ & $0.5\to0$ & $0.7\to0$\\
\midrule
Memory $(\lambda,k)$ & Vanilla / CFG & $(0,\mathrm{N/A})$ & $(0,\mathrm{N/A})$ & $(0,\mathrm{N/A})$\\
Memory $(\lambda,k)$ & PredMem & $(0.01,3)$ & $(0.3,3)$ & $(0.3,3)$\\
Memory $(\lambda,k)$ & CFG+PredMem & $(0.01,3)$ & $(0.3,3)$ & $(0.3,3)$\\
CFG $(w_S,w_X)$ & Vanilla / PredMem & disabled & disabled & disabled\\
CFG $(w_S,w_X)$ & CFG / CFG+PredMem & $(1.5,1.0)$ & $(1.5,1.0)$ & $(2.0,2.0)$\\
Read interface & PredMem / CFG+PredMem & post-LN attention & post-LN attention & token embedding\\
\bottomrule
\end{tabular}
\caption{\textbf{Motif-scaffolding settings.} Hyperparameters corresponding to Table~\ref{tab:motif}; fixed conditions are shared across arms.}
\label{tab:config_motif}
\end{table*}

\subsection{Structure Prediction}
\label{app:config_folding}
Table~\ref{tab:config_folding} lists the structure-prediction settings for Table~\ref{tab:full_conditional}, where the sequence is fixed and CFG is disabled. Because predictive memory operates across iterative refinement steps, DPLM-2 and DPLM-2.1 PredMem match the standard 100-step Default schedule (deterministic unmasking, zero temperature, and remasking enabled), whereas ESM3 PredMem matches its 8-step Vanilla schedule. ESM3 on CAMEO2022 has no PredMem entry (---) because its base decoding there uses a single step, leaving no iterative trajectory for predictive memory to act on. For completeness, we also report the native optimal-sampling Vanilla recipes ($L/8$ steps for DPLM-2 and 8 steps for DPLM-2.1).
\begin{table*}[t]
\centering\small
\setlength{\tabcolsep}{5pt}
\renewcommand{\arraystretch}{1.08}
\begin{tabular}{llccc}
\toprule
Setting & Arm / dataset & DPLM-2 & DPLM-2.1 & ESM3\\
\midrule
Steps & Default & 100 & 100 & 1\\
Unmasking & Default & deterministic & deterministic & deterministic\\
Temperature & Default & 0 & 0 & 0\\
Remasking & Default & enabled & enabled & disabled\\
\midrule
Steps & Vanilla & $L/8$ & 8 & 8\\
Unmasking & Vanilla & random & deterministic & stochastic\\
Temperature parameter & Vanilla & $0.1\to0$ & $0.1\to0$ & $0.5\to0$\\
Remasking & Vanilla & enabled & enabled & enabled\\
\midrule
Steps & PredMem & 100 & 100 & 8\\
Unmasking & PredMem & deterministic & deterministic & stochastic\\
Temperature parameter & PredMem & 0 & 0 & $0.5\to0$\\
Remasking & PredMem & enabled & enabled & enabled\\
Memory $(\lambda,k)$ & PredMem / CAMEO & $(0.5,1)$ & $(0.1,1)$ & ---\\
Memory $(\lambda,k)$ & PredMem / PDB & $(0.3,1)$ & $(0.03,3)$ & $(0.5,3)$\\
Memory $(\lambda,k)$ & Default / Vanilla & $(0,\mathrm{N/A})$ & $(0,\mathrm{N/A})$ & $(0,\mathrm{N/A})$\\
CFG & All & disabled & disabled & disabled\\
\bottomrule
\end{tabular}
\caption{\textbf{Structure-prediction settings.} Default and native optimal Vanilla recipes are reported alongside PredMem; a dash (---) marks a setting not evaluated.}
\label{tab:config_folding}
\end{table*}

\subsection{Inverse Folding}
\label{app:config_inverse}
Table~\ref{tab:config_inverse} lists the inverse-folding settings for Table~\ref{tab:full_conditional}, where the structure is fixed and only the sequence is generated. Across all three models, PredMem matches the exact step budget, unmasking strategy, temperature schedule, and remasking setting of the Default configuration (100-step deterministic decoding for DPLM-2/2.1; 8-step random unmasking with $1.0\to0$ annealing for ESM3), using $\lambda=0.5, k=1$ on both datasets for ESM3. The faster native Vanilla recipes (1 step for DPLM-2 and 8 steps for DPLM-2.1 and ESM3) are reported separately.
\begin{table*}[t]
\centering\small
\setlength{\tabcolsep}{5pt}
\renewcommand{\arraystretch}{1.08}
\begin{tabular}{llccc}
\toprule
Setting & Arm / dataset & DPLM-2 & DPLM-2.1 & ESM3\\
\midrule
Steps & Default & 100 & 100 & 8\\
Unmasking & Default & deterministic & deterministic & random\\
Temperature parameter & Default & 0 & 0 & $1.0\to0$\\
Remasking & Default & enabled & enabled & disabled\\
\midrule
Steps & Vanilla & 1 & 8 & 8\\
Unmasking & Vanilla & random & stochastic & deterministic\\
Temperature parameter & Vanilla & $0.1\to0$ & $0.1\to0$ & $0.5\to0$\\
Remasking & Vanilla & enabled & enabled & enabled\\
\midrule
Steps & PredMem & 100 & 100 & 8\\
Unmasking & PredMem & deterministic & deterministic & random\\
Temperature parameter & PredMem & 0 & 0 & $1.0\to0$\\
Remasking & PredMem & enabled & enabled & disabled\\
Memory $(\lambda,k)$ & PredMem / CAMEO & $(0.5,1)$ & $(0.01,3)$ & $(0.5,1)$\\
Memory $(\lambda,k)$ & PredMem / PDB & $(0.3,1)$ & $(0.5,1)$ & $(0.5,1)$\\
Memory $(\lambda,k)$ & Default / Vanilla & $(0,\mathrm{N/A})$ & $(0,\mathrm{N/A})$ & $(0,\mathrm{N/A})$\\
CFG & All & disabled & disabled & disabled\\
\bottomrule
\end{tabular}
\caption{\textbf{Inverse-folding settings.} Decoding and memory settings; PredMem matches the Default decoding schedule, while Vanilla lists the native optimal recipe.}
\label{tab:config_inverse}
\end{table*}

\subsection{Temperature, Position Selection, and Guidance}
\label{app:temperature}
For the DPLM annealing sampler,
\begin{equation}
 \tau_t=\tau_{\min}+(\tau_{\max}-\tau_{\min})(1-t/T),
 \label{eq:exp_dplm_temperature}
\end{equation}
where $t=0,\ldots,T-1$, so the final step remains slightly above the nominal lower endpoint. PredMem preserves all native position-selection and remasking rules.

For ESM3 multi-step annealing, the parameter $\theta$ is transformed as
\begin{equation}
 \tau_t=\left[\max\left(\theta-\frac{t}{\max(1,T-1)},0.001\right)\right]^2.
 \label{eq:exp_esm_temperature}
\end{equation}
Thus $\theta=0.5$ begins at an actual temperature of 0.25; over eight steps, the schedule takes values of approximately 0.25, 0.12755, 0.04592, 0.00510, followed by four steps at $10^{-6}$. Sampling temperatures are distinct from the memory-cache temperature, which is always fixed at one. Random position selection is confidence-agnostic, stochastic selection uses the native confidence-dependent randomized rule, and deterministic or entropy selection follows the native confidence ranking.

CFG and CFG+PredMem share identical guidance weights within each joint task. Guidance for unconditional co-generation drops the complementary modality in its condition-dropped branch, whereas motif guidance uses the native \texttt{drop\_motif\_cross\_modal} branch construction. Memory is written from each branch's raw logits prior to CFG, and caches remain strictly branch-local. In the primary conditional tasks, CFG is disabled so that the effect of predictive memory is evaluated in isolation.

\subsection{Memory Implementation and Computation}
\label{app:exp_memory}
\label{app:exp_compute}
All DPLM experiments use the post-LayerNorm attention read of Eq.~\eqref{eq:pm_postln} with residual-stream interpolation weight zero, whereas ESM3 uses token-embedding interpolation. Both generated tracks receive memory in joint tasks, and only the target track receives memory in conditional tasks (Appendix~\ref{app:pm_interfaces}).

Each forward pass updates the sample-, branch-, and position-local cache with a probability-weighted top-$k$ embedding mean. Top-$k$ selection, normalization, and aggregation are computed in FP32, and the detached cache is cast back to the native dtype at read time. While predictive summaries are cached across all positions, memory reads occur exclusively at eligible masked positions; fixed residues, supplied motif coordinates, and special tokens remain protected. The initial forward pass begins with an empty cache.

For $T$ steps and branch set $\mathcal G$, generation requires $T$ batched denoiser calls and $T|\mathcal G|$ branch evaluations (one branch for unguided decoding and three for joint CFG). PredMem preserves these exact evaluation counts with zero extra denoiser calls (NFE), returning a single terminal sample without test-time candidate search or reranking. The cache memory footprint is independent of $T$. Independent ESMFold refolding and optional terminal likelihood diagnostics are evaluation-only steps outside the generation budget.

\FloatBarrier
\section{Detailed Results}
\label{app:full_results}
\begin{table*}[!t]
\centering\scriptsize
\setlength{\tabcolsep}{2.5pt}
\renewcommand{\arraystretch}{1.12}
\begin{tabular}{@{}lllllll@{}}
\toprule
Model & Method & Success (\%) & pLDDT & scTM & scRMSD & Success clusters\\
\midrule
DPLM-2 & Vanilla & $29.80\pm2.76$ & $82.00\pm0.24$ & $0.852\pm0.0046$ & $4.95\pm0.21$ & $78.0\pm6.6$\\
DPLM-2 & PredMem & $33.40\pm1.89$ & $82.47\pm0.33$ & $0.852\pm0.0054$ & $4.79\pm0.21$ & $81.0\pm7.2$\\
DPLM-2 & CFG & $31.88\pm3.16$ & $82.42\pm0.29$ & $0.859\pm0.0049$ & $4.42\pm0.08$ & $73.4\pm6.5$\\
DPLM-2 & CFG+PredMem & $34.20\pm1.98$ & $82.99\pm0.44$ & $0.862\pm0.0063$ & $4.37\pm0.23$ & $74.6\pm3.9$\\
\midrule
DPLM-2.1 & Vanilla & $28.88\pm2.12$ & $84.91\pm0.44$ & $0.836\pm0.0017$ & $5.17\pm0.17$ & $92.6\pm3.6$\\
DPLM-2.1 & PredMem & $46.60\pm1.19$ & $87.77\pm0.27$ & $0.889\pm0.0032$ & $3.32\pm0.18$ & $75.6\pm5.9$\\
DPLM-2.1 & CFG & $32.88\pm0.72$ & $85.14\pm0.22$ & $0.853\pm0.0045$ & $4.37\pm0.22$ & $98.0\pm4.9$\\
DPLM-2.1 & CFG+PredMem & $48.32\pm2.05$ & $87.85\pm0.18$ & $0.895\pm0.0054$ & $3.13\pm0.16$ & $68.4\pm4.2$\\
\midrule
ESM3 & Vanilla & $33.36\pm1.08$ & $67.91\pm1.11$ & $0.545\pm0.0151$ & $31.91\pm0.94$ & $62.0\pm5.1$\\
ESM3 & PredMem & $82.08\pm1.01$ & $90.53\pm0.21$ & $0.944\pm0.0024$ & $2.32\pm0.18$ & $73.0\pm1.9$\\
ESM3 & CFG & $64.48\pm2.62$ & $88.36\pm0.71$ & $0.891\pm0.0113$ & $6.62\pm0.97$ & $114.4\pm2.3$\\
ESM3 & CFG+PredMem & $80.96\pm1.94$ & $91.59\pm0.22$ & $0.943\pm0.0022$ & $2.47\pm0.16$ & $84.4\pm0.9$\\
\bottomrule
\end{tabular}
\caption{\textbf{Unconditional co-generation results.} Full results corresponding to Table~\ref{tab:main}, reported as five-seed means $\pm$ SD.}
\label{tab:full_joint}

\par\medskip

\centering\scriptsize
\setlength{\tabcolsep}{2.5pt}
\renewcommand{\arraystretch}{1.12}
\begin{tabular}{@{}lllllllll@{}}
\toprule
Model & Method & Success (\%) & pLDDT & scTM & scRMSD & Motif RMSD & Solved /24 & Clusters\\
\midrule
DPLM-2 & Vanilla & $20.60\pm0.22$ & $72.21\pm0.12$ & $0.666\pm0.0019$ & $4.95\pm0.11$ & $2.54\pm0.04$ & $16.8\pm0.4$ & $37.8\pm2.8$\\
DPLM-2 & PredMem & $21.71\pm0.62$ & $72.23\pm0.15$ & $0.668\pm0.0025$ & $4.95\pm0.11$ & $2.53\pm0.04$ & $17.0\pm0.5$ & $41.0\pm4.8$\\
DPLM-2 & CFG & $20.48\pm0.81$ & $71.72\pm0.17$ & $0.661\pm0.0028$ & $5.03\pm0.08$ & $2.52\pm0.03$ & $16.2\pm0.8$ & $37.6\pm1.8$\\
DPLM-2 & CFG+PredMem & $21.25\pm0.78$ & $71.69\pm0.10$ & $0.669\pm0.0022$ & $4.95\pm0.07$ & $2.54\pm0.02$ & $16.4\pm0.5$ & $38.2\pm1.8$\\
\midrule
DPLM-2.1 & Vanilla & $18.33\pm0.80$ & $69.86\pm0.18$ & $0.659\pm0.0024$ & $5.38\pm0.05$ & $2.55\pm0.02$ & $15.2\pm0.8$ & $30.2\pm3.3$\\
DPLM-2.1 & PredMem & $20.29\pm0.49$ & $70.01\pm0.12$ & $0.671\pm0.0030$ & $5.16\pm0.04$ & $2.58\pm0.02$ & $15.4\pm0.5$ & $31.2\pm1.1$\\
DPLM-2.1 & CFG & $19.07\pm0.60$ & $68.47\pm0.09$ & $0.652\pm0.0032$ & $5.48\pm0.03$ & $2.49\pm0.03$ & $14.8\pm0.8$ & $30.0\pm2.5$\\
DPLM-2.1 & CFG+PredMem & $21.08\pm0.79$ & $69.09\pm0.18$ & $0.664\pm0.0041$ & $5.23\pm0.07$ & $2.56\pm0.21$ & $16.4\pm0.9$ & $36.8\pm4.3$\\
\midrule
ESM3 & Vanilla & $30.43\pm1.49$ & $76.25\pm0.52$ & $0.711\pm0.0066$ & $6.97\pm0.28$ & $0.61\pm0.06$ & $19.4\pm0.5$ & $57.4\pm4.4$\\
ESM3 & PredMem & $42.40\pm1.49$ & $77.89\pm0.34$ & $0.767\pm0.0053$ & $4.33\pm0.16$ & $0.73\pm0.10$ & $21.8\pm1.3$ & $64.8\pm5.4$\\
ESM3 & CFG & $35.81\pm1.29$ & $75.64\pm0.18$ & $0.754\pm0.0050$ & $4.29\pm0.17$ & $0.65\pm0.04$ & $22.6\pm0.5$ & $83.6\pm6.2$\\
ESM3 & CFG+PredMem & $41.93\pm1.31$ & $76.73\pm0.32$ & $0.776\pm0.0037$ & $3.75\pm0.17$ & $0.70\pm0.11$ & $22.0\pm1.0$ & $95.0\pm6.3$\\
\bottomrule
\end{tabular}
\caption{\textbf{Motif-scaffolding results.} Full results corresponding to Table~\ref{tab:motif}, with 2,400 outputs per run (100 samples for each of 24 motifs). Success requires scRMSD $<2$~\AA\ and motif RMSD $<1$~\AA, where motif RMSD compares the generator-decoded motif with the reference motif; pLDDT is ESMFold confidence on a 0--100 scale. Clusters count successful outputs within each motif, summed across motifs.}
\label{tab:full_motif}
\end{table*}

Tables~\ref{tab:full_joint} and~\ref{tab:full_motif} expand Tables~\ref{tab:main} and~\ref{tab:motif} in the same model and method order. Values report means $\pm$ sample standard deviations across five equally weighted runs. Zero standard deviation denotes identical recorded run-level values (e.g., under deterministic decoding) rather than an absence of variation across individual proteins. Rates are reported in percentages and RMSDs in~\AA.

\subsection{Unconditional Co-generation}
Success, pLDDT, scTM, scRMSD, and successful clusters correspond to Table~\ref{tab:main}.

\subsection{Motif Scaffolding}
The success criterion and within-motif clustering match Table~\ref{tab:motif}. The reported motif RMSD is measured on the generator-decoded structure because this quantity enters the success criterion directly.

\subsection{Structure Prediction and Inverse Folding}
\paragraph{Configurations and protocol.}
For single-modality conditional tasks, guidance is disabled so that memory feedback is evaluated in isolation. We report Default (the released reference configuration), Vanilla (the configured sampler following \citet{zhou2026unlocking}), and PredMem across CAMEO2022 (183 targets) and the PDB Date Split (449 targets). As detailed in Appendix~\ref{app:exp_protocol}, PredMem is paired with its schedule-matched multi-step baseline: Default for both DPLM models in both tasks (and for ESM3 in inverse folding), and Vanilla for ESM3 in structure prediction. All remaining configurations are retained in \cref{tab:full_conditional} as reference baselines. Structure prediction generates a structure from a fixed sequence and evaluates it against the reference using RMSD and TM-score, whereas inverse folding generates a sequence from a fixed backbone and measures refolded scTM and native amino-acid recovery (AAR).

\paragraph{Structure prediction.}
Against its schedule-matched baseline, PredMem lowers RMSD and raises TM-score on every model and dataset where it is evaluated (\cref{tab:full_conditional}). ESM3 on CAMEO2022 has no PredMem entry because its base decoding there is a single step, leaving no iterative trajectory for memory feedback. On ESM3, PDB RMSD decreases from $4.83$ to $4.27$~\AA, while TM-score improves over both the 8-step Vanilla baseline and the 1-step Default reference. DPLM-2 and DPLM-2.1 likewise improve over their Default schedules across both benchmarks, demonstrating consistent structural refinement even with a full sequence anchor.

\paragraph{Inverse folding.}
Relative to Default—which supplies PredMem's decoding schedule for all three models in this task—refolded scTM improves across all models and datasets (\cref{tab:full_conditional}), rising on ESM3 from $0.846$ to $0.854$ on CAMEO2022 and from $0.923$ to $0.930$ on PDB. ESM3 AAR also increases on both benchmarks ($46.51\%$ to $48.19\%$ on CAMEO2022; $49.62\%$ to $51.82\%$ on PDB). On DPLM-2 and DPLM-2.1, PredMem improves structural self-consistency (scTM) while keeping AAR closely aligned with the Default baseline.

\begin{table*}[!t]
\centering\scriptsize
\setlength{\tabcolsep}{2.5pt}
\renewcommand{\arraystretch}{1.12}
\begin{tabular}{@{}llllllllll@{}}
\toprule
& & \multicolumn{4}{c}{Structure prediction} & \multicolumn{4}{c}{Inverse folding}\\
\cmidrule(lr){3-6}\cmidrule(lr){7-10}
& & \multicolumn{2}{c}{CAMEO2022} & \multicolumn{2}{c}{PDB Date Split} & \multicolumn{2}{c}{CAMEO2022} & \multicolumn{2}{c}{PDB Date Split}\\
\cmidrule(lr){3-4}\cmidrule(lr){5-6}\cmidrule(lr){7-8}\cmidrule(lr){9-10}
Model & Method & RMSD & TM-score & RMSD & TM-score & scTM & AAR (\%) & scTM & AAR (\%)\\
\midrule
DPLM-2 & Default & $7.48\pm0.00$ & $0.686\pm0.0000$ & $5.26\pm0.00$ & $0.771\pm0.0000$ & $0.806\pm0.0031$ & $48.34\pm0.04$ & $0.875\pm0.0017$ & $55.59\pm0.09$\\
DPLM-2 & Vanilla & $7.45\pm0.07$ & $0.687\pm0.0025$ & $5.26\pm0.02$ & $0.769\pm0.0011$ & $0.808\pm0.0031$ & $48.32\pm0.04$ & $0.880\pm0.0017$ & $55.92\pm0.09$\\
DPLM-2 & PredMem & $7.34\pm0.00$ & $0.692\pm0.0000$ & $5.15\pm0.00$ & $0.773\pm0.0000$ & $0.808\pm0.0036$ & $48.19\pm0.07$ & $0.880\pm0.0020$ & $55.83\pm0.09$\\
\midrule
DPLM-2.1 & Default & $6.20\pm0.07$ & $0.747\pm0.0011$ & $2.79\pm0.02$ & $0.882\pm0.0006$ & $0.816\pm0.0018$ & $53.34\pm0.08$ & $0.901\pm0.0013$ & $78.25\pm0.03$\\
DPLM-2.1 & Vanilla & $6.13\pm0.05$ & $0.749\pm0.0005$ & $2.75\pm0.02$ & $0.883\pm0.0004$ & $0.820\pm0.0018$ & $53.30\pm0.08$ & $0.903\pm0.0013$ & $78.01\pm0.03$\\
DPLM-2.1 & PredMem & $6.12\pm0.06$ & $0.750\pm0.0007$ & $2.75\pm0.02$ & $0.883\pm0.0004$ & $0.819\pm0.0022$ & $53.33\pm0.06$ & $0.904\pm0.0011$ & $77.98\pm0.02$\\
\midrule
ESM3 & Default & $5.25\pm0.00$ & $0.794\pm0.0000$ & $4.02\pm0.00$ & $0.840\pm0.0000$ & $0.846\pm0.0027$ & $46.51\pm0.20$ & $0.923\pm0.0024$ & $49.62\pm0.06$\\
ESM3 & Vanilla & $5.60\pm0.10$ & $0.804\pm0.0017$ & $4.83\pm0.05$ & $0.842\pm0.0017$ & $0.846\pm0.0015$ & $48.51\pm0.05$ & $0.928\pm0.0008$ & $51.94\pm0.02$\\
ESM3 & PredMem & --- & --- & $4.27\pm0.04$ & $0.845\pm0.0002$ & $0.854\pm0.0010$ & $48.19\pm0.07$ & $0.930\pm0.0009$ & $51.82\pm0.01$\\
\bottomrule
\end{tabular}
\caption{\textbf{Structure prediction and inverse folding results.} Results on CAMEO2022 and PDB Date Split. Default is the released reference configuration and Vanilla is the configured sampler. RMSD is in~\AA; AAR is a percentage. Entries are means $\pm$ sample standard deviations across five runs. A dash (---) marks ESM3 on CAMEO2022, where single-step base decoding leaves PredMem no effect.}
\label{tab:full_conditional}
\end{table*}

\paragraph{Zero-condition structure generation.}
To contrast anchored and unanchored generation on the same target sets, \cref{tab:f0_self_vs_ref} evaluates structure-only generation when no sequence is revealed ($f=0\%$). Removing the sequence anchor yields substantially larger gains in internal self-consistency ($\Delta\text{scTM}=+0.4565$ on CAMEO2022 and $+0.3619$ on PDB for ESM3; $+0.0737$ on CAMEO2022 for DPLM-2.1) than in external-reference recovery, confirming that predictive memory provides its primary leverage in unanchored cross-modal regimes.

\begin{table*}[t]
\centering
\caption{\textbf{Self-consistency and reference recovery on zero-condition outputs.} Five-seed results on CAMEO2022 and PDB Date Split, with 20 targets per dataset and seed. Both metrics are evaluated on the same $f=0\%$ outputs, where $f$ is the revealed-sequence fraction (no sequence is revealed and only structure is generated).}
\label{tab:f0_self_vs_ref}
\footnotesize
\begin{tabular}{llccc}
\toprule
\textbf{Model} & \textbf{Dataset ($f=0\%$)} & \textbf{$\Delta\text{TM}_{\text{ref}}$ (External Ref.)} & \textbf{$\Delta\text{scTM}$ (Internal Self-Consistency)} & \textbf{Paired Gap $\Delta\text{scTM} - \Delta\text{TM}_{\text{ref}}$ [95\% CI]} \\
\midrule
\multirow{2}{*}{DPLM-2.1} & CAMEO2022 & $+0.0011 \pm 0.0082$ & $\mathbf{+0.0737 \pm 0.0496}$ & $\mathbf{+0.0727}\;\;[+0.0105,\;+0.1348]$ \\
 & PDB Date Split & $-0.0168 \pm 0.0246$ & $\mathbf{+0.0076 \pm 0.0175}$ & $\mathbf{+0.0244}\;\;[+0.0012,\;+0.0476]$ \\
\midrule
\multirow{2}{*}{ESM3} & CAMEO2022 & $+0.0900 \pm 0.0163$ & $\mathbf{+0.4565 \pm 0.0583}$ & $\mathbf{+0.3665}\;\;[+0.3105,\;+0.4225]$ \\
 & PDB Date Split & $+0.0818 \pm 0.0250$ & $\mathbf{+0.3619 \pm 0.0709}$ & $\mathbf{+0.2801}\;\;[+0.2138,\;+0.3464]$ \\
\bottomrule
\end{tabular}
\end{table*}

\begin{figure*}[t]
\centering
\includegraphics[width=\textwidth]{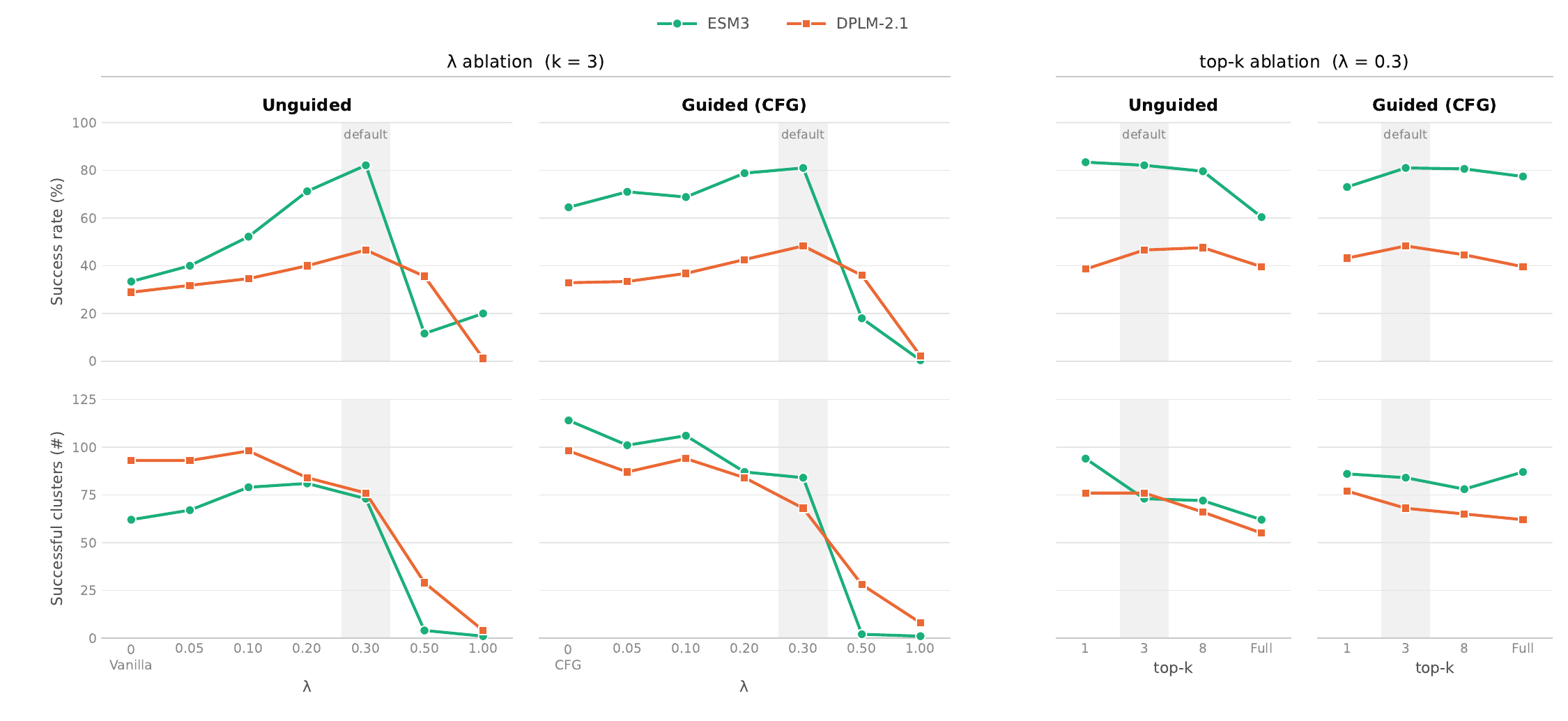}
\caption{\textbf{Mixing-weight and support-size ablations.} Left: $\lambda$ ablation at $k{=}3$; right: top-$k$ ablation at $\lambda{=}0.3$. Top row: success rate; bottom row: number of successful clusters. Each panel compares ESM3 and DPLM-2.1 under unguided and guided (CFG) decoding. Shaded bands mark the default setting; $\lambda{=}0$ corresponds to Vanilla (unguided) or CFG (guided).}
\label{fig:analysis_ablation}
\end{figure*}

\subsection{Mixing Weight, Support Size, and Read Interface}
\label{sec:analysis_ablation}
\label{sec:analysis_modality}
\label{sec:modality}
\label{app:analysis_modality}
\label{app:analysis_strength}

\paragraph{Moderate mixing weights perform best.}
With dual-stream PredMem and $k{=}3$, we sweep the interpolation weight $\lambda$ (left half of Figure~\ref{fig:analysis_ablation}). Success rises with $\lambda$ and peaks at the default $\lambda{=}0.3$ on both backbones under both unguided and guided decoding (ESM3: $33.4\%\to82.1\%$ unguided and $64.5\%\to81.0\%$ guided; DPLM-2.1: $28.9\%\to46.6\%$ and $32.9\%\to48.3\%$ from $\lambda{=}0$). Overly aggressive interpolation degrades generation: $\lambda{=}0.5$ already reduces ESM3 to $11.6\%$ (unguided) and $18.0\%$ (guided), and $\lambda{=}1.0$ drives DPLM-2.1 to $1.2\%$ and $2.2\%$ and guided ESM3 to $0.4\%$, because excessive feedback overrides the native \texttt{MASK} representation that the frozen denoiser expects at unresolved sites.

\paragraph{Successful clusters.}
Successful-cluster counts (bottom row) reveal the same quality--coverage trade-off. Under guided decoding, increasing $\lambda$ from $0$ (CFG) to $0.3$ reduces successful clusters from $114$ to $84$ on ESM3 and from $98$ to $68$ on DPLM-2.1 while raising success, and coverage collapses at larger weights (guided ESM3 retains only $2$ clusters at $\lambda{=}0.5$). Unguided decoding differs across backbones: ESM3 gains clusters at the default weight ($62$ to $73$) because far more outputs become successful, whereas DPLM-2.1 trades $93$ clusters for $76$.

\paragraph{Support size.}
Holding $\lambda{=}0.3$, we vary the support size $k$ (right half of Figure~\ref{fig:analysis_ablation}). Under guided decoding, $k{=}3$ achieves the highest success on both backbones ($81.0\%$ on ESM3, $48.3\%$ on DPLM-2.1), whereas expanding to the full vocabulary lowers success to $77.4\%$ and $39.6\%$. Without guidance, ESM3 remains near its optimum for $k\leq8$ ($83.4\%$, $82.1\%$, and $79.6\%$ for $k{=}1,3,8$) before dropping to $60.4\%$ at full support, while DPLM-2.1 peaks at $k{=}8$ ($47.6\%$) with $k{=}3$ within $1.0$~pp. Larger supports also tend to reduce coverage: unguided successful clusters decrease from $94$ to $62$ on ESM3 and from $76$ to $55$ on DPLM-2.1 as $k$ grows from $1$ to full, and guided DPLM-2.1 decreases from $77$ to $62$ (while guided ESM3 stays within $78$--$87$). Because the default $k{=}3$ lies within $1.3$~pp of the best success in every setting, $k$ acts as a smooth knob trading off peak success against structural breadth.

\paragraph{Read interface.}
Table~\ref{tab:preln_vs_postln} compares replacing the Post-LN read with a Pre-LN read under identical read-weight settings on unguided unconditional co-generation. On DPLM-2, success drops from $33.40\%$ to $0.44\%$ and clusters fall from $81.0$ to $1.4$; on DPLM-2.1, success drops from $46.60\%$ to $28.0\%$ and clusters from $75.6$ to $3.0$. These results align with the normalization-cancellation effect analyzed in Appendix~\ref{app:pm_dplm}.

\begin{table*}[t]
\centering\small
\caption{\textbf{Read-interface ablation for PredMem on unguided unconditional co-generation.} Both Post-LN (from Table~\ref{tab:full_joint}) and Pre-LN entries report five-seed means $\pm$ SD under the same read-weight settings. Replacing Post-LN with Pre-LN severely degrades co-generation success and structural coverage, confirming the normalization-cancellation effect in Appendix~\ref{app:pm_dplm}.}
\label{tab:preln_vs_postln}
\begin{tabular}{llccccc}
\toprule
Model & Read interface & Success (\%) $\uparrow$ & pLDDT $\uparrow$ & scTM $\uparrow$ & scRMSD (\AA) $\downarrow$ & Clusters $\uparrow$ \\
\midrule
\multirow{2}{*}{DPLM-2}
 & Pre-LN  & $0.44 \pm 0.36$  & $53.35 \pm 0.71$ & $0.330 \pm 0.0033$ & $27.24 \pm 0.51$ & $1.4 \pm 1.1$ \\
 & Post-LN & $\mathbf{33.40 \pm 1.89}$ & $\mathbf{82.47 \pm 0.33}$ & $\mathbf{0.852 \pm 0.0054}$ & $\mathbf{4.79 \pm 0.21}$  & $\mathbf{81.0 \pm 7.2}$ \\
\midrule
\multirow{2}{*}{DPLM-2.1}
 & Pre-LN  & $28.0 \pm 7.60$   & $80.07 \pm 0.96$ & $0.871 \pm 0.0022$ & $\mathbf{2.95 \pm 0.16}$  & $3.0 \pm 0.0$ \\
 & Post-LN & $\mathbf{46.60 \pm 1.19}$ & $\mathbf{87.77 \pm 0.27}$ & $\mathbf{0.889 \pm 0.0032}$ & $3.32 \pm 0.18$  & $\mathbf{75.6 \pm 5.9}$ \\
\bottomrule
\end{tabular}
\end{table*}

\section{Temperature, Entropy, and Overhead \texorpdfstring{\mbox{Analyses}}{Analyses}}
\label{app:analysis_controls}
This appendix expands Section~\ref{sec:analysis} with complete temperature sweeps, entropy-floor and positional controls, and runtime and memory overhead measurements. Each analysis uses a matched baseline, sample budget, and seed set.

\begin{figure*}[t]
\centering
\includegraphics[width=0.92\textwidth]{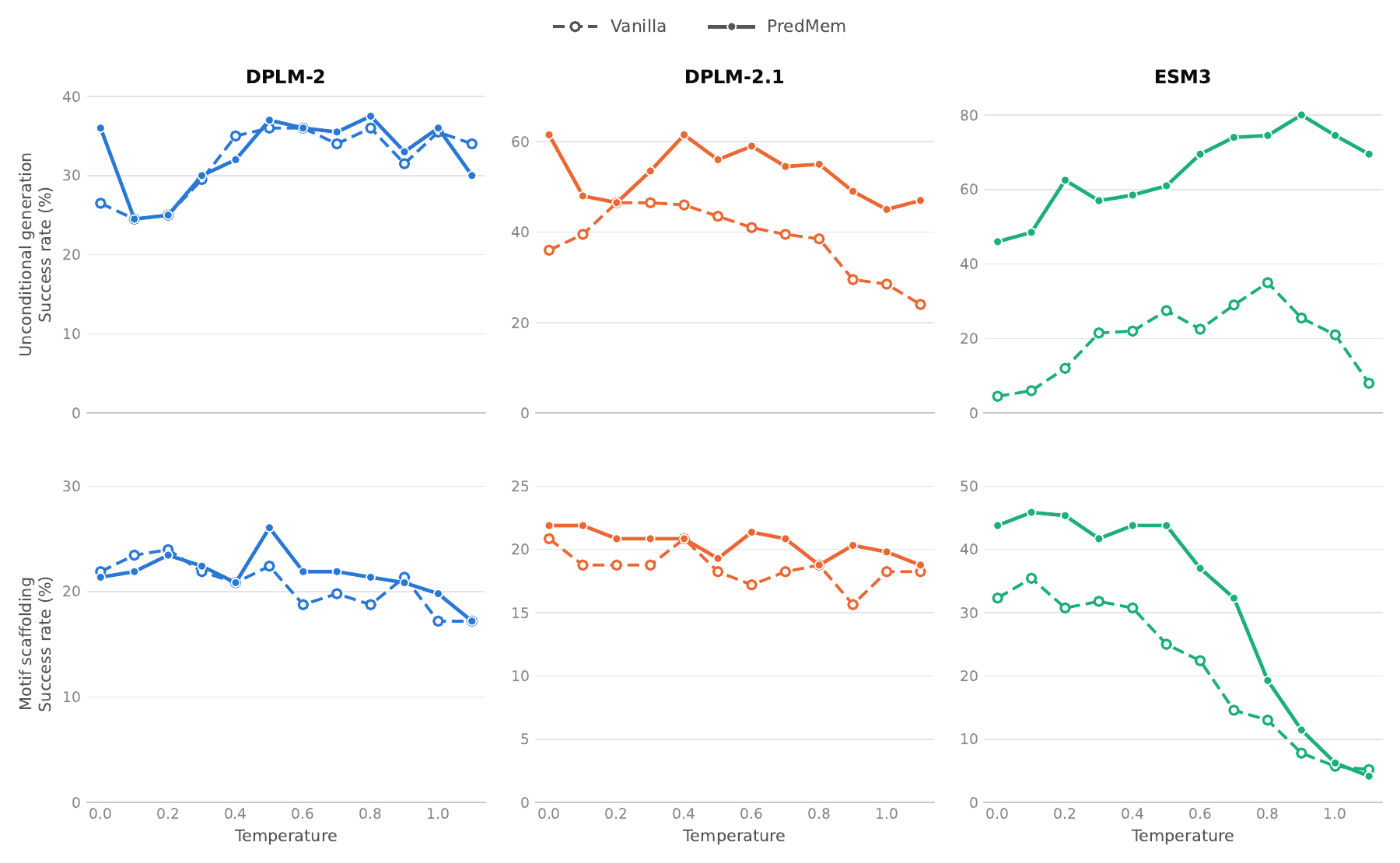}
\caption{\textbf{Complete temperature sweep.} Top: unconditional co-generation; bottom: motif scaffolding. Columns: DPLM-2, DPLM-2.1, ESM3. Dashed lines with open markers show Vanilla and solid lines with filled markers show PredMem, each over 12 constant temperatures without CFG (200 unconditional or 192 motif outputs per point from a single run).}
\label{fig:analysis_temperature_all}
\end{figure*}

\subsection{Complete Constant-Temperature Sweep}
\label{app:analysis_temperature}
We evaluate three models, two tasks, two methods, and 12 constant sampling temperatures $\tau\in\{0.0,0.1,\ldots,1.1\}$ (144 configurations and 28,224 outputs from a single run, with CFG disabled). Unconditional co-generation generates 40 outputs at each length $L\in\{100,200,300,400,500\}$ (200 per configuration), while motif scaffolding generates 8 outputs for each of 24 variants (192 per configuration). Within each model--task pair, decoding order, remasking, step budget, and PredMem settings are held fixed across the temperature grid and follow Tables~\ref{tab:config_joint} and~\ref{tab:config_motif}, while the memory-cache temperature remains one. These constant-temperature runs complement the annealed main-table recipes in Appendix~\ref{app:exp_protocol}.

Figure~\ref{fig:analysis_temperature_all} shows that PredMem's gains persist across sampling temperatures rather than relying on a single tuned operating point. On ESM3, PredMem outperforms Vanilla at all 12 temperatures in unconditional co-generation (+45.00~pp peak gain) and 11 of 12 temperatures in motif scaffolding (+10.42~pp peak gain), with every unconditional PredMem point exceeding Vanilla's global maximum. On DPLM-2.1, PredMem wins 11 of 12 temperatures in unconditional co-generation (+15.00~pp peak gain) and matches or exceeds Vanilla at all 12 temperatures in motif scaffolding (10 wins, 2 ties; +1.04~pp peak gain). On DPLM-2, where unguided baselines are already more stable across temperatures, PredMem improves grid maxima by +1.50~pp in unconditional co-generation (7 wins, 3 ties, 2 losses) and by +2.08~pp in motif scaffolding.
\FloatBarrier

\subsection{Entropy Floor and Content Controls}
\label{app:analysis_entropy}
To rule out simple distribution sharpening as the primary driver of PredMem's gains, we compare against a same-state entropy-floor control (PredMem+floor in Table~\ref{tab:counterfactual_controls_full}). At each PredMem-visited hard state, let $g_0$ and $g_M$ denote the no-memory reference logits and PredMem logits for a given position and modality. With original sampling temperature $\tau$, we define the target entropy floor
\begin{align}
 h^\star&=\max\{H(\operatorname{softmax}(g_0/\tau)),\nonumber\\[-2pt]
 &\hspace{48pt}H(\operatorname{softmax}(g_M/\tau))\},\label{eq:analysis_entropy_target}\\
 H(p)&=-\sum_v p(v)\log p(v),\nonumber
\end{align}
and solve for a per-position inverse-temperature factor $0<\beta\leq1$ such that $H(\operatorname{softmax}(\beta g_M/\tau))=h^\star$ (keeping $\beta=1$ whenever PredMem already meets or exceeds the reference entropy). The controller bisects for 48 iterations in float64 and verifies float32 sampling entropy to within $2\times10^{-4}$~nats, raising the effective temperature $\tau/\beta\geq\tau$ while preserving PredMem's top-1 identity and token ranking.

Under unguided decoding (Table~\ref{tab:counterfactual_controls_full}), PredMem+floor retains about $80\%$ of PredMem's gain over Vanilla on ESM3 ($72.5\%$ vs.~$82.1\%$ for PredMem and $33.4\%$ for Vanilla) and about $40\%$ on DPLM-2.1 ($36.0\%$ vs.~$46.6\%$ and $28.9\%$). Entropy reduction therefore accounts for a small part of the gain on ESM3 and a larger part on DPLM-2.1, while a clear improvement over Vanilla remains on both.

Tables~\ref{tab:counterfactual_controls_full} and~\ref{tab:temp_fitting} provide the full numerical breakdown for the unguided counterfactual controls and same-state temperature-fitting diagnostics summarized in Section~\ref{sec:analysis}. In Table~\ref{tab:temp_fitting}, even after fitting an optimal positive scalar temperature to align Vanilla with PredMem at identical states, 64.76\%--99.56\% of the KL divergence remains unexplained and Top-1 predictions change in every setting (2.57\%--94.28\%), confirming that PredMem reshapes token rankings rather than merely rescaling temperature. The magnitude varies across models and stages: DPLM-2.1 alters up to 94.28\% of Top-1 predictions early and ESM3 alters up to 62.87\% at the Mid-early stage, whereas DPLM-2 and DPLM-2.1 change at most 6.38\% at the Mid-early stage (KL below 0.004~nats).

\begin{table}[t]
\centering
\caption{\textbf{Controls under unguided decoding.} Five-seed means $\pm$ SD across seeds at the sample budget of Table~\ref{tab:main}; $\Delta$ columns give percentage-point differences within each model. Vanilla $2T$ doubles the steps. PredMem+shift cyclically reassigns memory among unresolved positions by $\lfloor n/2\rfloor$ in residue order, separately per stream. PredMem+floor raises PredMem's token entropy to at least the no-memory reference at the same state while preserving token ranking.}
\label{tab:counterfactual_controls_full}
\footnotesize
\setlength{\tabcolsep}{3pt}
\begin{tabular}{llccc}
\toprule
\textbf{Model} & \textbf{Arm} & \textbf{Success (\%)} & \shortstack{\textbf{$\Delta$ vs.}\\\textbf{Vanilla}} & \shortstack{\textbf{$\Delta$ vs.}\\\textbf{PredMem}} \\
\midrule
\multirow{5}{*}{ESM3} & Vanilla & $33.4 \pm 1.1$ & --- & $-48.7$ \\
 & Vanilla~$2T$ & $34.5 \pm 1.2$ & $+1.1$ & $-47.6$ \\
 & PredMem+shift & $48.8 \pm 1.2$ & $+15.4$ & $-33.3$ \\
 & PredMem+floor & $72.5 \pm 2.0$ & $+39.1$ & $-9.6$ \\
 & \textbf{PredMem} & $\mathbf{82.1 \pm 1.0}$ & $+48.7$ & --- \\
\midrule
\multirow{5}{*}{DPLM-2.1} & Vanilla & $28.9 \pm 2.1$ & --- & $-17.7$ \\
 & Vanilla~$2T$ & $30.4 \pm 5.2$ & $+1.5$ & $-16.2$ \\
 & PredMem+shift & $35.1 \pm 6.1$ & $+6.2$ & $-11.5$ \\
 & PredMem+floor & $36.0 \pm 6.0$ & $+7.1$ & $-10.6$ \\
 & \textbf{PredMem} & $\mathbf{46.6 \pm 1.2}$ & $+17.7$ & --- \\
\bottomrule
\end{tabular}
\end{table}

\begin{table}[t]
\centering
\caption{\textbf{Same-state temperature fitting under unguided decoding.} At identical decoding states, a positive scalar temperature $T$ is fitted per position to minimize $\mathrm{KL}(p^{\text{PredMem}}\parallel\operatorname{softmax}(g^{\text{Vanilla}}/T))$. For DPLM-2.1's factorized 13-bit structure head, a single shared scalar $T$ is fitted across all 13 binary heads at each position. Full KL is measured at $T{=}1$ and reports the mean across positions, and the unexplained ratio reports the per-position median residual ratio. \mbox{Top-1} changes cannot be produced by positive scalar scaling. Mid-early denotes roughly 25\% of decoding progress. Structure vocabularies: 13-bit (DPLM-2.1), 8192-way (DPLM-2), 4096-way (ESM3).}
\label{tab:temp_fitting}
\footnotesize
\setlength{\tabcolsep}{2.5pt}
\begin{tabular}{llccc}
\toprule
\textbf{Model} & \textbf{Stage / Modality} & \shortstack{\textbf{Top-1}\\\textbf{changed (\%)}} & \shortstack{\textbf{Full KL}\\\textbf{(nats)}} & \shortstack{\textbf{Unexpl.}\\\textbf{ratio (\%)}} \\
\midrule
\multirow{4}{*}{DPLM-2.1} & Early / Sequence & 63.74 & 0.1721 & 84.38 \\
 & Early / Structure & 94.28 & 0.4541 & 79.43 \\
 & Mid-early / Sequence & 2.57 & 0.0034 & 64.76 \\
 & Mid-early / Structure & 6.38 & 0.0038 & 90.63 \\
\midrule
\multirow{4}{*}{DPLM-2} & Early / Sequence & 25.45 & 0.0104 & 96.79 \\
 & Early / Structure & 29.62 & 0.0207 & 98.49 \\
 & Mid-early / Sequence & 2.93 & 0.0026 & 78.76 \\
 & Mid-early / Structure & 3.04 & 0.0012 & 95.12 \\
\midrule
\multirow{4}{*}{ESM3} & Early / Sequence & 8.56 & 0.0003 & 89.79 \\
 & Early / Structure & 40.60 & 0.0492 & 99.56 \\
 & Mid-early / Sequence & 17.26 & 0.1380 & 74.15 \\
 & Mid-early / Structure & 62.87 & 0.5969 & 91.33 \\
\bottomrule
\end{tabular}
\end{table}

\subsection{Runtime and Memory Overhead}
\label{app:analysis_overhead}
Table~\ref{tab:overhead} reports the cost of PredMem relative to the same decoder without memory at each sequence length. Without guidance, PredMem adds 3.0--3.2\% decoding time on ESM3 and 12.4--13.2\% on DPLM-2.1—consistent with the 3.1\% (ESM3) and 12.8\% (DPLM-2.1) average overheads reported in Section~\ref{sec:analysis}—with at most $0.007$~GiB and $0.026$~GiB of extra VRAM, respectively. Under CFG, the overhead is 3.8--4.5\% on ESM3 and 8.9--13.2\% on DPLM-2.1, with at most $0.014$~GiB and $0.12$~GiB of extra VRAM (where the $0.12$~GiB peak was measured at batch size $10$).

\begin{center}
\begin{minipage}{\columnwidth}
\small
\refstepcounter{table}\label{tab:overhead}
\noindent Table~\thetable. \textbf{Runtime and VRAM overhead of PredMem.} For each sequence length $L$, Time gives the increase in decoding time and VRAM the additional GPU memory in GiB, for PredMem relative to Vanilla (unguided) and for CFG+PredMem relative to CFG (guided).
\par\vspace{4pt}
\centering\footnotesize
\setlength{\tabcolsep}{2.5pt}
\begin{tabular}{llrrrr}
\toprule
 &  & \multicolumn{2}{c}{PredMem / Vanilla} & \multicolumn{2}{c}{CFG+PredMem / CFG} \\
\cmidrule(lr){3-4}\cmidrule(lr){5-6}
Model & $L$ & \multicolumn{1}{c}{Time (\%)} & \multicolumn{1}{c}{VRAM (GiB)} & \multicolumn{1}{c}{Time (\%)} & \multicolumn{1}{c}{VRAM (GiB)} \\
\midrule
\multirow{5}{*}{ESM3} & 100 & 3.16 & 0.0012 & 3.84 & 0.0023 \\
 & 200 & 3.17 & 0.0024 & 3.85 & 0.0049 \\
 & 300 & 3.03 & 0.0039 & 3.87 & 0.0078 \\
 & 400 & 3.14 & 0.0049 & 4.13 & 0.0098 \\
 & 500 & 3.12 & 0.0068 & 4.47 & 0.0137 \\
\midrule
\multirow{5}{*}{DPLM-2.1} & 100 & 12.69 & 0.0224 & 12.40 & 0.0265 \\
 & 200 & 12.36 & 0.0234 & 13.23 & 0.0322 \\
 & 300 & 12.89 & 0.0263 & 12.76 & 0.0298 \\
 & 400 & 13.24 & 0.0249 & 10.76 & 0.0462 \\
 & 500 & 13.19 & 0.0243 & 8.87 & 0.0489 \\
\bottomrule
\end{tabular}
\end{minipage}
\end{center}

\FloatBarrier

\section{Code Availability}
\label{app:code}
Code is available at \url{https://github.com/XL-S224/PredMem}

\FloatBarrier

\end{document}